\documentclass{article}
\usepackage{graphicx} 
\usepackage[utf8]{inputenc} 
\usepackage[T1]{fontenc}    
\usepackage{hyperref}       
\usepackage{url}            
\usepackage{booktabs}       
\usepackage{amsfonts}       
\usepackage{nicefrac}       
\usepackage{microtype}      
\usepackage{xcolor}         
\usepackage{algorithm}
\usepackage{algpseudocode}
\usepackage{subfigure}
\usepackage{arydshln}
\usepackage{makecell}
\usepackage{tabularray}
\usepackage[normalem]{ulem}
\usepackage{amsmath}
\usepackage{amssymb}
\usepackage{mathtools}
\usepackage{amsthm}
\usepackage{float}
\usepackage[many]{tcolorbox}
\tcbuselibrary{listings,skins}
\usepackage{multicol}
\usepackage{multirow}
\usepackage{longtable}
\usepackage{verbatim}
\usepackage{enumitem}
\usepackage{natbib}
\usepackage[final]{style}
\usepackage{listings}
\usepackage{xcolor}
\usepackage{soul}
\usepackage{ulem}
\usepackage{caption}

\theoremstyle{plain}

\theoremstyle{definition}

\theoremstyle{remark}

\title{HeurAgenix: Leveraging LLMs for Solving Complex Combinatorial Optimization Challenges}
\author{
  Xianliang Yang$^{1}$ \quad
  Ling Zhang$^{1}$ \quad
  Haolong Qian$^{1,2}$ \quad
  Lei Song$^{1}$ \quad
  Jiang Bian$^{1}$ \\
  $^{1}$Microsoft Research Asia, Beijing, China \\
  $^{2}$Tsinghua University, Beijing, China \\
  \texttt{\{Xianliang.Yang, Ling.Zhang, v-haolqian, Lei.Song, Jiang.Bian\}} \\
  \texttt{@microsoft.com}
}
\date{May 2025}

\begin{document}

\maketitle

\begin{abstract}
Heuristic algorithms play a vital role in solving combinatorial optimization (CO) problems, yet traditional designs depend heavily on manual expertise and struggle to generalize across diverse instances. We introduce \textbf{HeurAgenix}, a two-stage hyper-heuristic framework powered by large language models (LLMs) that first evolves heuristics and then selects among them automatically. In the heuristic evolution phase, HeurAgenix leverages an LLM to compare seed heuristic solutions with higher-quality solutions and extract reusable evolution strategies. During problem solving, it dynamically picks the most promising heuristic for each problem state, guided by the LLM's perception ability. For flexibility, this selector can be either a state-of-the-art LLM or a fine-tuned lightweight model with lower inference cost. To mitigate the scarcity of reliable supervision caused by CO complexity, we fine-tune the lightweight heuristic selector with a dual-reward mechanism that jointly exploits singals from selection preferences and state perception, enabling robust selection under noisy annotations. Extensive experiments on canonical benchmarks show that HeurAgenix not only outperforms existing LLM-based hyper-heuristics but also matches or exceeds specialized solvers. Code is available at \url{https://github.com/microsoft/HeurAgenix}.
\end{abstract}
\section{Introduction}
\label{sec:introduction}

Combinatorial optimization (CO) problems are fundamental in operations research and critical for decision-making across various industries~\cite{cook1994combinatorial, nemhauser1998integer}. These problems often involve large-scale search spaces, where dimensionality grows exponentially, making traditional solution methods computationally intractable~\cite{papadimitriou1998combinatorial, ausiello2012complexity}. This complexity has led to the widespread use of heuristics, which provide approximate solutions within a feasible time frame~\cite{power2007model, schulze2016energy}. Heuristics are typically designed to be interpretable, with clear rule-based decision-making processes that enhance transparency and human understanding.

Despite their effectiveness, heuristics heavily rely on manual expertise and are challenging to adapt to changing conditions. In response, hyper-heuristics have emerged, aiming to automate heuristic design by crafting rules for their selection and combination~\cite{burke2003hyper, burke2013hyper}. Although hyper-heuristics offer interpretability, they often require manual rule design, which limits adaptability to evolving problem states~\cite{peres2021combinatorial, branke2015automated}.

Recent advancements in large language models (LLMs) have inspired work on automatic heuristic design, using LLMs to generate and refine heuristics through few-shot prompting and code synthesis~\cite{romera2024mathematical, liu2024evolution,ye2024reevo,novikov2025alphaevolve}. Although these approaches attain strong performance on moderate-sized classic CO problems, most of them embed the LLM-produced heuristic inside a task-specific solver, which resulting reliance on hand-crafted domain knowledge constrains heuristic flexibility and the generalization.

To address these limitations, we introduce \textbf{HeurAgenix}, a unified framework for automatic heuristic evolution and adaptive selection. To the best of our knowledge, HeurAgenix is the first LLM-based hyper-heuristic framework that simultaneously (i) evolves a diverse pool of heuristics without relying on any external solver and (ii) incorporates an online heuristic selector for adaptive problem solving. Depending on user requirements, the second phase can leverage either an frontier LLM (such as GPT~\cite{achiam2023gpt}, DeepSeek~\cite{deepseek2025}) or a fine-tuned lightweight model for efficient inference. A diagram illustrating the proposed framework is provided in Figure~\ref{fig:framework}. A diagram illustrating the proposed framework is provided in Figure~\ref{fig:framework}. Our main contributions are:

\begin{figure}[t]
    \begin{center}
    \centerline{\includegraphics[scale=0.9]{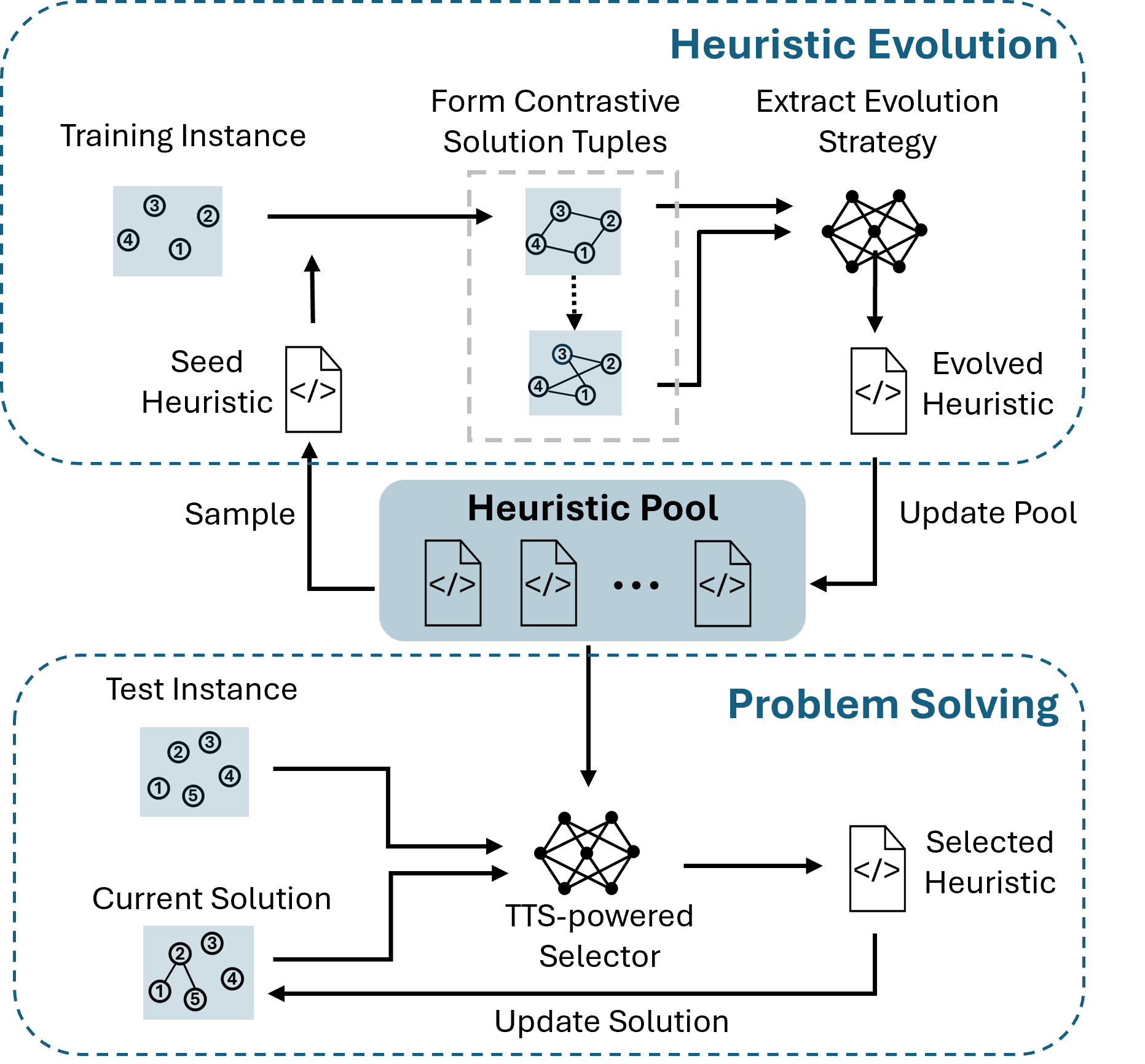}}
    \caption{Overview of the HeurAgenix framework for automatic heuristic design and adaptive selection. In the heuristic evolution phase, an LLM autonomously discovers evolution strategies by analyzing contrastive solution tuples, while in the problem solving phase, an adaptive heuristic selection mechanism integrates Test-time Scaling (TTS)~\cite{wei2022chain,wang2022self}.}
    \label{fig:framework}
    \end{center}
    \vskip -0.1in
\end{figure}

\begin{itemize}
    \item We introduce \textbf{HeurAgenix}, a versatile framework for automatic heuristic evolution and selection. It offers scalable and generalizable solutions for complex CO problems, outperforming existing hyper-heuristic approaches.
    \item We propose a \textbf{contrastive, data-driven heuristic evolution} phase, analyzing solution trajectories to discover evolution strategies without predefined rules.
    We develop an \textbf{adaptive heuristic selection mechanism} that integrates LLM and TTS, which selects heuristics based on the current problem state and improve efficiency and solution quality.
    \item To train the heuristic selector more robustly under noisy data, we introduce a \textbf{dual-reward mechanism} that combines context perception for accurate state recognition with outcome preference that amplifies the positive/negative margin.
\end{itemize}
\section{Preliminary and Related Work}
\label{sec:preliminary_and_related_work}
In this section, we present the important definitions and notations employed throughout this paper.

\subsection{Heuristics for CO Problems}
\label{sec:heuristic_definition}
\paragraph{Problem state} In CO problems, directly characterizing a problem instance and its solution can be challenging, as they are often represented in complex numerical data structures. To address this, following the CO community, we adopt the concept of a \textbf{problem state} as an abstract representation capturing the high-level features of both the problem instance and its current (partial) solution~\cite{burke2009survey}. For example, in the Traveling Salesman Problem (TSP), static problem states may include features such as the number of nodes, average inter-node distance, and graph symmetry, while dynamic problem states can describe aspects such as visited nodes and current tour cost.

\paragraph{Heuristic} In this paper, we define a heuristic algorithm as a function that maps a \textbf{problem state} \(z\) to an \textbf{operation} \(O\). Formally, a heuristic \(H\) can be described as \(H: \mathcal{Z} \rightarrow \mathcal{O}\), where \(\mathcal{Z}\) represents the space of problem states and \(\mathcal{O}\) is the set of allowable operations.  If the heuristic is a \textbf{constructive heuristic}, the operation \(O\) may involve adding elements to extend a partial solution. Conversely, if the heuristic is an \textbf{improvement heuristic}, \(O\) might involve exchanging, replacing, or perturbing existing elements to refine the solution~\cite{gigerenzer2011heuristic}. Detailed designs for heuristics, problem states and operations are provided in the Appendix~\ref{sec:detailed_parameter_settings}.

\paragraph{Transition function.}  
For a problem instance \(d\) we first define the single-step \textbf{transition function}  
\[
\mathcal T:\;\mathcal Z \times \mathcal O \;\longrightarrow\; \mathcal Z,
\]  
which deterministically maps the current state and an applied operation to the next state.  For later convenience we extend the definition of \(\mathcal T\) to accept a heuristic as its second argument:  
\[
\mathcal T(z, H)\;:=\;\mathcal T\!\bigl(z,\,H(z)\bigr),
\]  
that is, query the heuristic for its chosen operation at \(z\) and then perform the original state update.  Repeatedly executing the same heuristic \(H\) for exactly \(M\) steps is denoted  
\[
\mathcal T^{M}(z, H) \;=\; 
\underbrace{\mathcal T\bigl(\dotsm\mathcal T(\mathcal T(z,H),H),H\bigr)}_{\text{$M$ successive applications of $H$}},
\]  
a shorthand we will use extensively in Section~\ref{sec:problem_solving} when defining the selector’s optimization target.

\paragraph{Solution trajectory.}  
Given a sequence of heuristics \((H_0,\dots,H_{n-1})\), the resulting trajectory is  
\[
z_0 = \mathrm{Init}(d),\;
O_0 = H_0(z_0),\;
z_1 = \mathcal T(z_0,O_0),\;
\dots,\;
O_{n-1} = H_{n-1}(z_{n-1}),\;
z_n = \mathcal T(z_{n-1},O_{n-1}),
\]  
where \(z_i\) is the state before step \(i\) and \(O_i\) the selected operation.  We write the trajectory as \(S \;=\;\{(z_i,O_i)\}_{i=0}^{n-1}\) and denote its objective cost by \(C(S)\).

\paragraph{Heuristic selector.}
Given the problem state \(z \in \mathcal{Z}\), the remaining decisions \(t\in\{0,\dots, T\}\), and the heuristic pool \(\mathcal{H}\), we define a heuristic selector as a mapping  
\[
\pi:\;\mathcal{Z}\times\{0,\dots,T\}\;\longrightarrow\;\mathcal{H},
\]  
which selects the heuristic \(\pi(z,t)\) at problem state \(z\) when \(t\) decisions remain.

\subsection{Hyper-Heuristics}
\label{sec:hyper_heuristics}
In the CO community, hyper-heuristics have been introduced to manipulate heuristics. Two main categories are generation hyper-heuristics and selection hyper-heuristics~\cite{burke2009survey}.

\textbf{Generation hyper-heuristics} involve the automatic creation of heuristics by systematically combining elementary operations or decision-making rules. These techniques often employ methods such as genetic programming, genetic algorithms, and particle swarm optimization~\cite{qing2023generalize, singh2022study}. Although these methods can yield high-performing algorithms, they often encounter challenges related to computational overhead and adaptability~\cite{wu2021learning, jia2019taso}. 

\textbf{Selection hyper-heuristics} dynamically choose the most appropriate heuristic from a predefined set by evaluating the current problem state. These methods incorporate rule-based, meta-heuristic, or learning-based strategies, rendering them effective for complex optimization tasks. Nevertheless, they may struggle with intricate selection mechanisms and generalization issues~\cite{drake2020recent, de2021comparative, de2020hyper, sopov2016selection}.

\subsection{LLMs for Combinatorial Optimization}
\label{sec:llms_for_combinatorial_optimization}

\begin{table}[h]
    \caption{Comparison of LLM-based CO paradigms.}
    \label{tab:comparison_between_LLM_based_heuristics}
    \resizebox{\columnwidth}{!}{
    \begin{tabular}{llll}
    \toprule
    Paradigm & Heuristic Evolution & Problem Solving & Solver Required$^{\ast}$ \\
    \midrule
    FunSearch & LLM-driven & Fixed heuristic & Yes \\
    EoH & 5 manually designed strategies & Fixed heuristic & Yes \\
    ReEvo & Feedback-based refinement & Fixed heuristic & Yes \\
    AlphaEvolve & Ensemble LLM-driven evolution & Fixed heuristic & No \\
    HeurAgenix (Ours) & Contrastive, data-driven evolution & Adaptive selection & No\\ 
    \bottomrule
    \end{tabular}}
    \vskip 0.1in
    $^{\ast}$A solver refers to either:
    a traditional optimization solver (e.g., GLS~\cite{voudouris2010guided}), 
    a neural network-based solver (e.g., PoMo~\cite{kwon2020pomo}), 
    or a specialized hyper-heuristic algorithm (e.g., ACO~\cite{dorigo2007ant}).
\end{table}

LLMs have demonstrated significant potential in addressing combinatorial optimization (CO) problems. For instance, Zhang et al.~\cite{zhang2024can} assessed the performance of LLMs on various graph optimization challenges, while Iklassov et al.~\cite{iklassov2024self} developed effective prompting strategies to enable LLMs to adapt to diverse problem formulations. Xiao et al.~\cite{xiao2023chain} introduced the Chain-of-Experts approach, integrating multi-agent cooperation to directly address optimization tasks. These studies underscore LLMs' flexibility and capability in reasoning within CO problems.

Particularly relevant to our work are studies focusing on LLMs for heuristic generation and evolution in CO problem solving. FunSearch~\cite{romera2024mathematical} employs LLMs to iteratively generate and refine candidate solutions, aiming to improve solution quality. EoH~\cite{liu2024evolution} facilitates multi-directional evolution to boost heuristic diversity, while ReEvo~\cite{ye2024reevo} leverages LLM-driven reflection for targeted optimization. Collectively, these efforts highlight LLMs' potential in automating heuristic design and enhancing optimization processes. Most recently, AlphaEvolve~\cite{novikov2025alphaevolve} pushes the paradigm further by pairing an ensemble of Gemini~\cite{team2023gemini} LLMs with automated evaluators in an evolutionary loop, enabling the discovery of entire algorithmic codebases that optimise practical systems.

However, as illustrated in Table~\ref{tab:comparison_between_LLM_based_heuristics}, methods such as FunSearch, EoH, and ReEvo couple the LLM-generated heuristic with a task-specific solver, which limits generalization and demands domain expertise. AlphaEvolve removes the external solver by evolving complete programs, yet its offline evolutionary loop is computationally intensive and does not provide instance-level adaptation at inference time. In contrast, HeurAgenix supports flexible real-time heuristic selection by leveraging either an LLM or a fine-tuned lightweight model. This enables autonomous data-driven evolution and on-the-fly adaptation among multiple heuristics without dependency on external solvers, thus forming a truly end-to-end optimization paradigm.

\subsection{Test-time Scaling}
\label{sec:test_time_scaling}

Test-time scaling (TTS) is the practice of intentionally spending extra computation at inference time to improve answer quality~\cite{graves2016adaptive,figurnov2017spatially,teerapittayanon2016branchynet}.  Early research demonstrated a compute–accuracy trade-off in progressive or interruptible inference; recent LLM work has turned TTS into a standard tool for eliciting stronger reasoning, for example by sampling multiple candidate outputs and selecting the best one with a lightweight verifier~\cite{wei2022chain,wang2022self}. Representative instantiations include Best-of-$N$ sampling~\cite{brown2024largelanguagemonkeysscaling} and Diverse-Verifier Tree Search (DVTS) \cite{chen2024llmcallsneedscaling}. Our framework adopts the same search-verifier philosophy in problem solving phase to trade a modest increase in inference compute for substantial gains in solution quality and robustness.

\subsection{Challenges in Noisy Data for Long-Term Decision-Making}
\label{sec:challenges_in_noisy_data_for_long_sequence_decision-making}

Training models for effective long-term decision making is fundamentally challenged by the difficulty of accurately assessing the true long-term value of each action (i.e., selecting a particular heuristic at a given step). This difficulty stems from the credit-assignment problem: the ultimate impact of a single selection may only emerge after many steps or even after the whole trajectory, which complicates isolating its precise contribution, especially within large state–action spaces and over extended horizons. Exhaustively evaluating the long-term utility of early selections would require enumerating every possible continuation, which is computationally infeasible~\cite{noisylabelsrevisited}.

As a result, any practical evaluation procedure can provide only approximate, noisy scores based on a small subset of rollouts. Training models on such imperfect feedback therefore demands robust learning techniques. HeurAgenix meets this need via a dual-reward mechanism that fuses final-solution preferences with intermediate-reasoning signals, yielding more reliable supervision for heuristic selection under noisy annotations.
\section{Methodology}
\label{sec:methodology}

HeurAgenix is designed to leverage the complementary strengths of generation and selection hyper-heuristics, offering a robust framework for combinatorial optimization. 

As illustrated in Figure~\ref{fig:framework}, the generative component utilizes the reasoning abilities of LLMs to autonomously refine and evolve heuristic algorithms across diverse problem instances. Meanwhile, the selection component enhances adaptability by dynamically choosing the most suitable heuristic based on the current problem state, employing a fine-tuned lightweight model. By unifying these methods, HeurAgenix achieves high performance and adaptability, effectively addressing the challenges of large-scale CO problems.

\subsection{Heuristic Evolution}
\label{sec:heuristic_evolution}

Heuristic evolution is a systematic, data-driven procedure that iteratively improves a \textbf{seed heuristic} $H_\text{seed}$.  We leverage an LLM both to diagnose structural weaknesses and to propose concrete evolution strategies that amend them.

Let \(\mathcal{D}_{\mathrm{evo}}\) and \(\mathcal{D}_{\mathrm{val}}\) denote the evolution and validation instance sets, respectively. Building on the definitions in Section~\ref{sec:heuristic_definition}, a single evolution round proceeds as follows:

\textbf{Step 1: Basic solution generation.}
For every $d\!\in\!\mathcal{D}_{\mathrm{evo}}$  we run $H_\text{seed}$ to obtain a \textbf{basic solution} $S=\{(z_i,O_i)\}_{i=0}^{n-1}$ and its cost $C(S)$. If $H_\text{seed}$ is a constructive heuristic, it is applied repeatedly from the initial state until a feasible complete solution is constructed. If $H_\text{seed}$ is an improvement heuristic, we first sample a constructive heuristic $H_c$ to obtain an initial feasible solution, then apply $H_\text{seed}$ iteratively as a local search or improvement operator.

\textbf{Step 2: Contrastive solution generation.}
To reveal potential weaknesses, we perform up to $P$ perturbation trials.  In each trial we
(i) randomly select a subset $\mathcal{K} \subset \{0,\ldots,n-1\}$ of operation indices in the basic solution,
(ii) replace every $O_k\,(k\!\in\!\mathcal {K})$ by an alternative $O'_k\!\in\!\mathcal{O}(z_k)\setminus\{O_k\}$,
(iii) roll out the modified trajectory to obtain a perturbed solution $S'$ and its cost $C(S')$.
Whenever $C(S')<C(S)$ (for minimization problems), $S'$ is stored as a \textbf{contrastive solution}; its mutation list $\mathcal{M}=\{(z_k,O_k,O'_k)\}_{k\in\mathcal K}$ becomes a candidate operation set for evolution.

\textbf{Step 3: Critical operation identification.}
Not all modified operations in $\mathcal{M}$ are responsible for the performance gap, so we first identify the decisive mutation. For each $(z_k,O_k,O'_k)\in\mathcal{M}$ we independently replace $O_k$ with $O'_k$, keep all other operations still from $H_{seed}$, and if the solution are valid, measure the individual improvement
\[
\Delta_k = C(S) - C\!\bigl(S^{(k)}\bigr),
\]
where $S^{(k)}$ is the solution obtained after changing only the $k$-th operation (positive $\Delta_k$ denotes improvement for a minimization task). We then select
$k^\star=\arg\max_{k\in\mathcal{K}}\Delta_k$
and \(O_{k^\star}\) is the \textbf{critical operation}.

\textbf{Step 4: Extract evolution strategy.}
The tuple $(z_{k^\star},O_{k^\star},O'_{k^\star})$ and the current heuristic $H_\text{seed}$ are passed to the LLM:
\[
\mathcal{E} \;=\; \operatorname{LLM}_{\text{evolve}}\bigl(H_\text{seed},\,z_{k^\star},\,O_{k^\star},\,O'_{k^\star}\bigr).
\]
The LLM explains why $O'_{k^\star}$ is better and outputs an \textbf{evolution strategy} $\mathcal{E}$. This strategy serves as a conceptual guide to improve $H_\text{seed}$, potentially involving modifications such as parameter adjustment, the addition of control logic, or component replacement.

\begin{figure}[h]
    \centering
    \includegraphics[width=\linewidth]{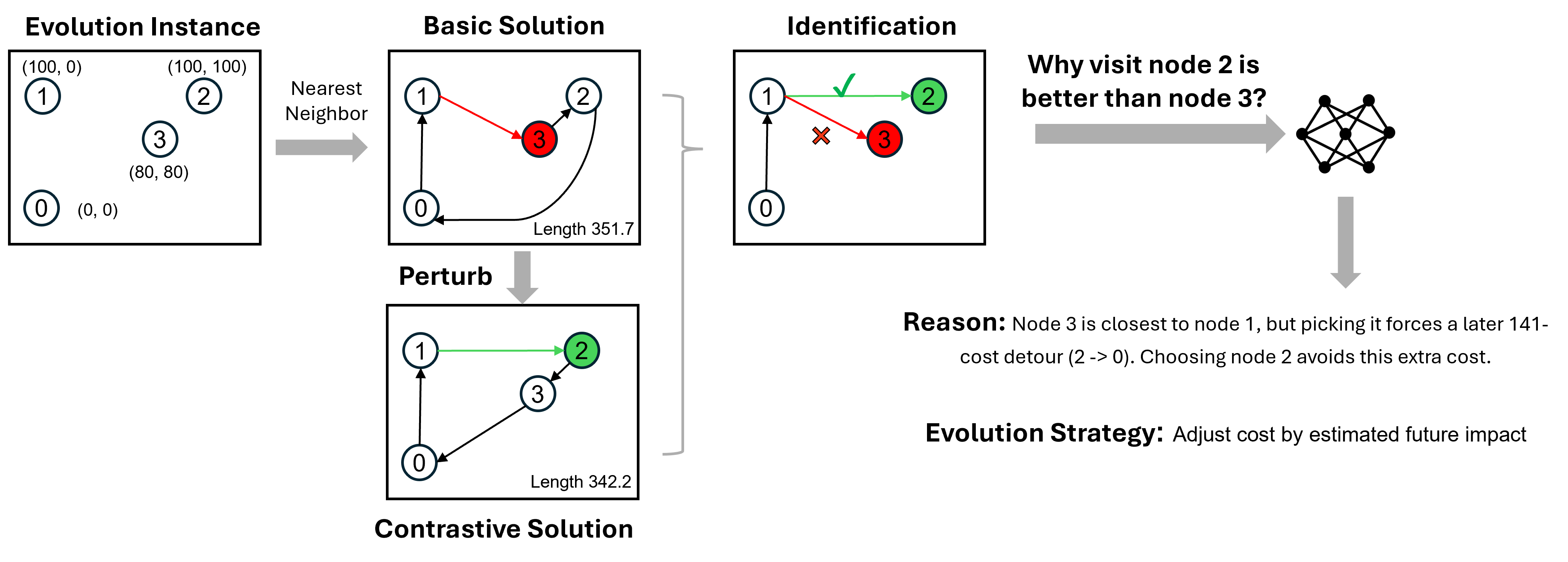}
    \caption{Illustration of one heuristic-evolution step on a four-node TSP evolution instance. The cumulative effect of this and subsequent refinements can be seen in Figure~\ref{fig:evolution_example}.}
    \label{fig:single_evolution_example}
    \vskip -0.1in
\end{figure}

\textbf{Step 5: Iterative refinement.}
Starting from $H_0\!=\!H_\text{seed}$, we refine the heuristic for at most $T_{\max}$ rounds.
After each round we compute the performance
$
p_i \;=\; \operatorname{evaluate\_performance}\bigl(H_i,\mathcal{D}_{\mathrm{val}}\bigr)
      \;=\;\frac{1}{|\mathcal{D}_{\mathrm{val}}|}
        \sum_{d\in\mathcal{D}_{\mathrm{val}}}C\!\bigl(S^i_d\bigr)
$, where $S^i_d$ denotes the solution produced by $H_i$ on instance $d$, 
and obtain an update
$
H_{i+1}=\operatorname{LLM}_{\text{refine}}\bigl(H_i,\mathcal{E},p_i\bigr).
$
We stop early if no further improvement is observed.
Algorithm~\ref{alg:one_round_evolution} summarizes the entire procedure. Figure~\ref{fig:single_evolution_example} also visualizes how a single critical operation mutation is extracted and justified.

\begin{algorithm}
\caption{One Round Heuristic Evolution}
\label{alg:one_round_evolution}
\textbf{Input:} seed heuristic $H_\text{seed}$; evolution instance $d$; validation set $\mathcal{D}_{\mathrm{val}}$; LLM; maximum perturbation trials $P$; maximum refinement trials $I_{\max}$ \\
\textbf{Output:} refined heuristic $H_{\text{evolved}}$
\begin{algorithmic}[1]
    \footnotesize
    \State \textcolor{gray}{\# Step 1: Basic Solution Generation}
    \State $S \gets \mathrm{run\_heuristic}(H_\text{seed}, d)$ \Comment{$S=\{(z_i,O_i)\}_{i=0}^{n-1}$}
    \State $n \gets |S|$
    \Statex
    \State \textcolor{gray}{\# Step 2: Contrastive Solution Generation}
    \State $\mathcal{M} \gets \varnothing$;\quad $p \gets 0$
    \While{$(p < P)\ \land\ (\mathcal{M} = \varnothing)$}
        \State $p \gets p + 1$
        \State $\mathcal{K} \gets \mathrm{sample\_indices}(n)$ \Comment{a random subset of operation indices}
        \State $S' \gets \mathrm{perturb}(S,\mathcal{K})$ \Comment{replace each $O_k$ with $O'_k$}
        \If{$C(S') < C(S)$} \Comment{for a minimization problem}
            \State $\mathcal{M} \gets \{(z_k,O_k,O'_k)\}_{k\in\mathcal{K}}$
        \EndIf
    \EndWhile
    \If{$\mathcal{M} = \varnothing$}
        \State \Return $H_\text{seed}$ \Comment{no better contrastive solution found}
    \EndIf
    \Statex
    \State \textcolor{gray}{\# Step 3: Critical Operation Identification}
    \ForAll{$(z_k,O_k,O'_k) \in \mathcal{M}$}
        \State $S^{(k)} \gets \mathrm{perturb\_single}(S,k,O'_k)$
        \State $\Delta_k \gets C(S) - C(S^{(k)})$
    \EndFor
    \State $k^\star \gets \arg\max_{k}\,\Delta_k$
    \State $(z_{k^\star},O_{k^\star},O'_{k^\star}) \gets$ tuple with index $k^\star$
    \Statex
    \State \textcolor{gray}{\# Step 4: Extract Evolution Strategy}
    \State $\mathcal{E} \gets \operatorname{LLM}_{\text{evolve}}(H_\text{seed}, z_{k^\star}, O_{k^\star}, O'_{k^\star})$
    \Statex
    \State \textcolor{gray}{\# Step 5: Iterative Heuristic Refinement}
    \State $t \gets 0$;\quad $H_0 \gets H_\text{seed}$;\quad \textit{improved} $\gets$ \textbf{true}
    \While{$\textit{improved} \land i < I_{\max}$}
        \State $p_i \gets \text{evaluate\_performance}(H_i,\; \mathcal{D}_{\mathrm{val}})$
        \State $H_{i+1} \gets \operatorname{LLM}_{\text{refine}}(H_i,\; \mathcal{E},\; p_i)$
        \State $p_{i+1} \gets \text{evaluate\_performance}(H_{i+1},\; \mathcal{D}_{\mathrm{val}})$
        \If{$p_{i+1} < p_i$}  \Comment{for a minimization problem}
            \State $i \gets i + 1$
        \Else
            \State \textit{improved} $\gets$ \textbf{false}
        \EndIf
    \EndWhile
    \State $H_{\text{evolved}} \gets H_i$
    \State \Return $H_{\text{evolved}}$
\end{algorithmic}
\end{algorithm}

\begin{figure}[h]
    \centering
    \includegraphics[width=\linewidth]{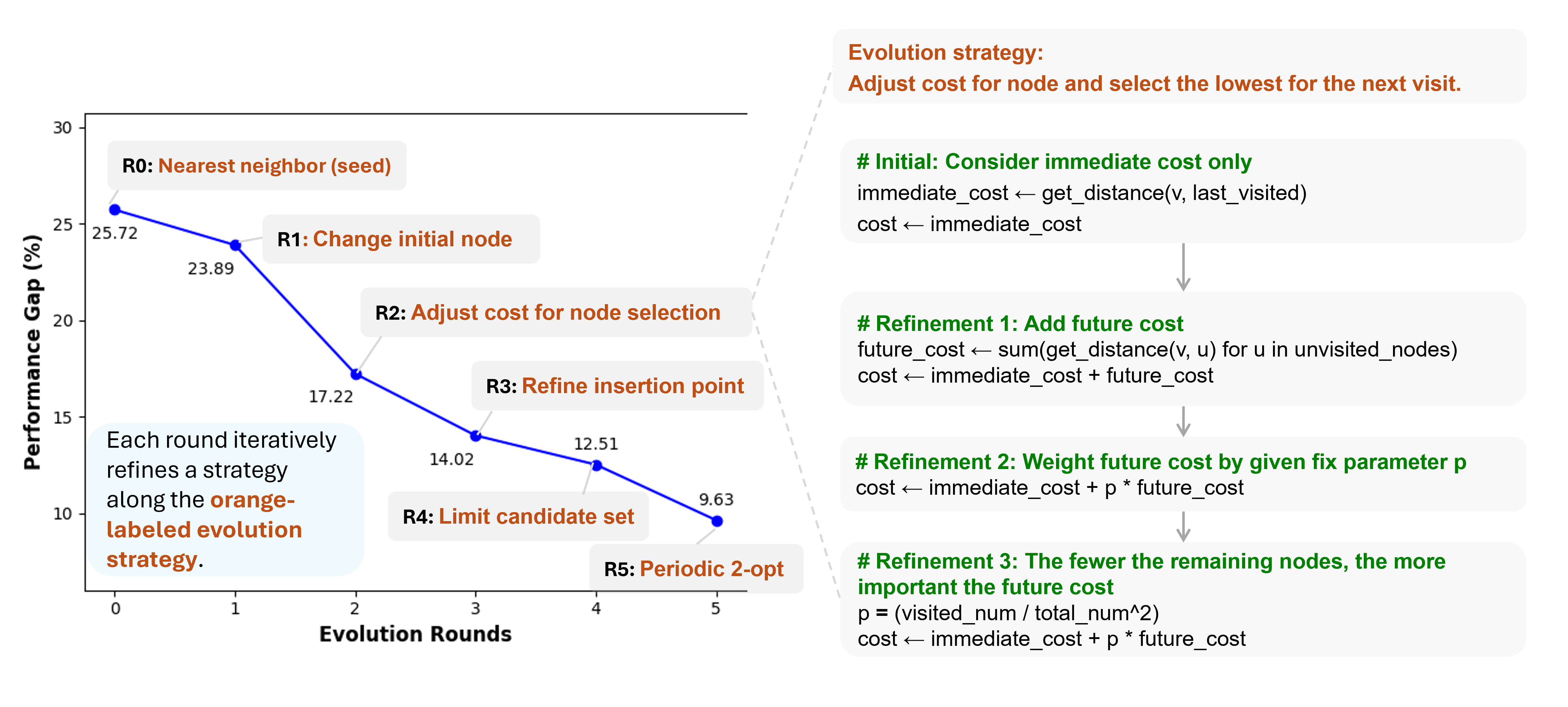}
    \caption{Example of heuristic evolution for TSP. The left panel illustrates successive strategy refinements and their impact on TSPLIB~\cite{reinelt1991tsplib} performance. The right panel details a specific evolution step, where an alternative cost function is induced by the LLM based on counterfactual analysis. For a step-by-step extraction of the refinement highlighted in Round 2, see Figure~\ref{fig:single_evolution_example} and for further details and code, see Appendix~\ref{sec:heuristic_evolution_example}.}
    \label{fig:evolution_example}
    \vskip -0.1in
\end{figure}

\paragraph{Algorithmic Summary and Discussion.}
Each round of evolution enhances the heuristic in a data-driven manner. By repeating the process over diverse evolution instances and aggregating the learned strategies, the heuristic pool $\mathcal{H}$ grows in diversity and effectiveness. Figure~\ref{fig:evolution_example} gives an overview of multiple rounds of evolution and their corresponding performance changes. For further details on an evolution example and hyper-parameter choices, refer to Appendix~\ref{sec:heuristic_evolution_example} and Appendix~\ref{sec:detailed_parameter_settings}.

\subsection{Problem Solving}
\label{sec:problem_solving}

\paragraph{Motivation.} Even after evolution, a single heuristic rarely dominates across all problem states; performance varies sharply with the current state.  Static, hand-crafted switching rules therefore lack adaptability~\cite{burke2006case, guan2021differential, kim2019appliance}. We therefore require a real-time selector that decides which heuristic should be applied next for each encountered state \(z\) (defined as Section~\ref{sec:heuristic_definition}).
\paragraph{Heuristic selection objective.}
Following the notation of Section~\ref{sec:heuristic_definition}, for a minimization problem we jointly define the state-value function, the action value of a heuristic, and the corresponding heuristic selector:
\begin{equation}\label{eq:heuristic_selector_objector}
\begin{aligned}
V(z,t) &=
\begin{cases}
C(S_z), & t = 0,\\[4pt]
\displaystyle \min_{H\in\mathcal H}
        V\!\bigl(\mathcal T^{M}(z,H),\,t-1\bigr), & \bigl\lceil\tfrac{N}{M}\bigr\rceil \ge t > 0,
\end{cases}\\[8pt]
Q(z,H,t) &= V\!\bigl(\mathcal T^{M}(z,H),\,t-1\bigr),\\[6pt]
\pi^\star(z,t) &= \arg\min_{H\in\mathcal H} Q(z,H,t),
\end{aligned}
\end{equation}
where
\begin{itemize}
    \item \(N\) is the maximum number of heuristic calls, 
    \item \(M\) is the fixed number of consecutive steps per chosen heuristic, 
    \item \(t\) represents remaining decisions (where \(t = 0\) means termination), 
    \item \(S_z\) is the current solution corresponding to state \(z\), and 
    \item \(\mathcal H\) denotes the evolved heuristic pool.
\end{itemize} 
From these, \(V(z,t)\) represents the expected cost from state \(z\) with \(t\) decisions remaining, \(Q(z,H,t)\) evaluates a heuristic's performance in that state, and \(\pi^\star(z,t)\) represents the optimal selector that chooses heuristics with minimum \(Q\). The objective of heuristic selector optimization is to determine the optimal selector. For simplicity, we will omit the parameter \(t\) if it is clear from the context.

\paragraph{Heuristic selection procedure.}
In practice, obtaining $Q(z,H)$ exactly is infeasible. Monte-Carlo simulation is a common surrogate, but exhaustively simulating every heuristic in $\mathcal H$ is prohibitively slow on large instances, whereas an LLM—relying on coarse semantic cues alone—typically misses the best choice.  
We therefore adopt a hybrid strategy: the LLM first prunes the pool to a compact candidate set, after which a lightweight test-time search (TTS) evaluates those candidates and selects the one to execute:

\textbf{Step 1: Filter candidate heuristics.}
The LLM filters the heuristic pool to produce a candidate subset
\(\mathcal H'=\operatorname{LLM}_\text{filter}(z, \mathcal H)\).

\textbf{Step 2: Evaluate value of heuristics.}
For every \(H\in\mathcal H'\) under problem state $z$, we estimate its value by Monte-Carlo search~\cite{swiechowski2023monte}: \( \hat{Q_H} = \operatorname{mc\_evaluate}(z, H) \). Specifically, we first apply it \(H\) sequentially \(M\) times starting from state \(z\), resulting in a new intermediate state. From the intermediate state, we then generate \(T\) candidate solutions by repeatedly selecting heuristics randomly from \(\mathcal{H}\) until completion. The estimated value $\hat{Q_H}$ is the average terminal cost over the \(T\) rollouts—averaging provides an unbiased estimate of the expected outcome, while taking a minimum would introduce a downward bias and favor lucky samples. Implementation details appear in Appendix~\ref{sec:monte_carlo_evaluation_strategy}.

\textbf{Step 3: Select the most suitable heuristic.}
Select  
\(\hat H=\arg\min_{H\in\mathcal H'} \hat{Q_H}\)  
and execute it \(M\) times.

This structured approach couples rapid LLM-based reasoning with a lightweight search verifier, yielding a balanced and efficient solver that leverages both model guidance and TTS.

\paragraph{Problem solving process.} Based on this heuristic selector, we can dynamic select the heuristic and solve problem easily. For a test instance \(d\) we proceed iteratively.  Starting from the initial state \(z\), the selector chooses a heuristic $\hat H$, and then execute $\hat H$ is executed \(M\) times, to update the problem . The cycle repeats until none of the available heuristics can further improve the solution.  The hyper-parameter \(M\) balances decision frequency against runtime overhead and is set to \(M=5\) throughout all experiments.

\subsection{Selection Model Fine-tuning}
\label{sec:selection_model_fine_tuning}

\paragraph{Motivation} Selecting the optimal heuristic in real-time plays a crucial role in combinatorial optimization performance. To enhance efficiency, we propose the implementation of lightweight selection models within HeurAgenix to effectively manage latency, computational demands, and resource utilization. This approach is particularly beneficial as it allows for multiple heuristic evaluations and selections without incurring prohibitive inference costs associated with larger models.

Lightweight models face inherent limitations in processing complex problem states and applying various heuristics effectively, particularly in their ability to adaptively select appropriate heuristics for different scenarios. To address these challenges, we propose a fine-tuning approach that enhances the model's comprehension of both the problem domain and available heuristics. Our methodology begins with an offline data collection process, followed by a detailed analysis of our implementation strategy for extracting insights from non-optimal and noisy datasets.

\paragraph{Offline Data Collection}
We construct an offline dataset consisting of tuples \((z,\;H,\;Q_H)\). Here \(z\) is the current problem state, \(H\) is the candidate heuristic drawn from the pool \(\mathcal {H}\), and \(Q_H\) is the rollout value assigned to \(H\) at state \(z\). To obtain \(Q_H\), every heuristic in \(\mathcal {H}\) is evaluated from the initial state \(z\) by the Monte Carlo pipeline of selection, expansion, simulation, and back-propagation (see Appendix \ref{sec:monte_carlo_evaluation_strategy}), yielding a scalar score for that \((z, H)\) pair.

To obtain data that is both informative and robust, we collect two types of trajectories. 
(i) Greedy trajectories update the state with the heuristic that attains the highest \(Q_H\) at each step, exposing the selector to near-optimal decisions. 
(ii) Stochastic trajectories update the state with a random heuristic, forcing the selector to recover from non-optimal contexts. 
The resulting mixture encourages the model to associate diverse states with effective heuristics.

As noted in Section~\ref{sec:challenges_in_noisy_data_for_long_sequence_decision-making}, the rollout values \(Q_H\) are inherently noisy because only a small subset of future continuations can be explored. 
We therefore employ the dual-reward fine-tuning scheme to learn policies that remain reliable under such imperfect supervisory signals.

\begin{figure}[!htbp]
  \centering
  \includegraphics[width=1.0\textwidth]{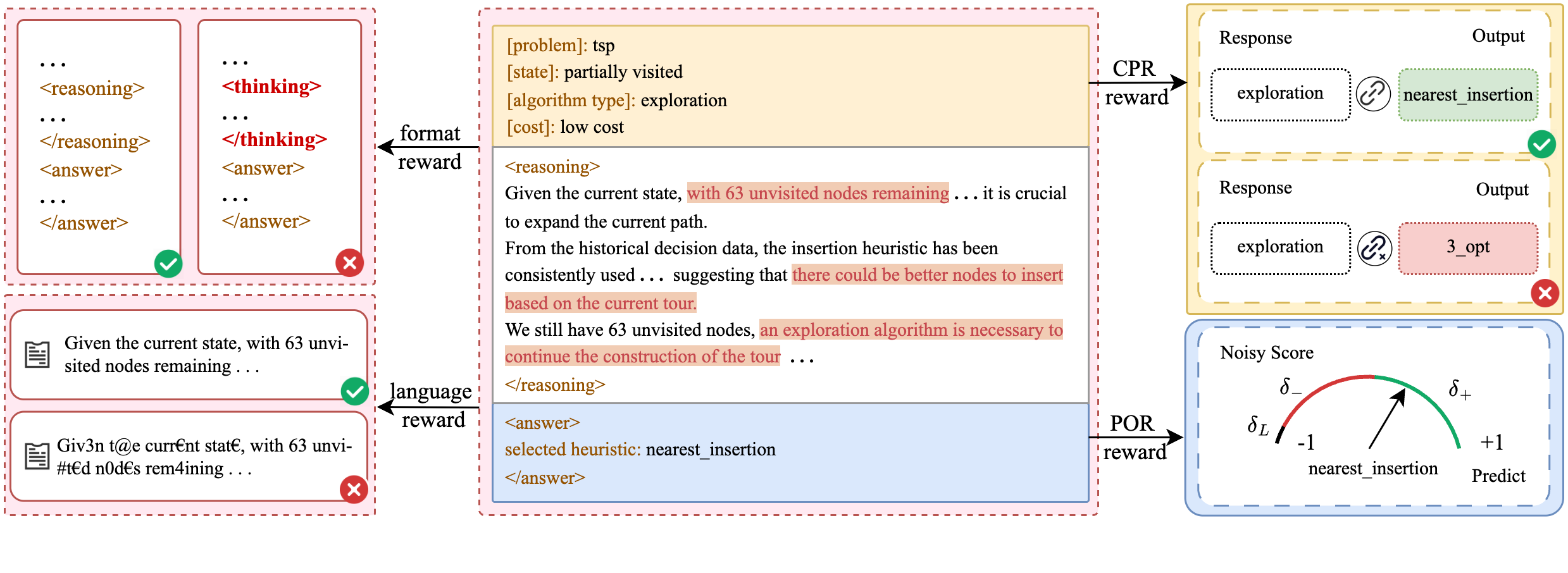}
  \caption{Detailed reward design, showing the operational mechanisms of the novel POR and CPR as well as auxiliary Format Reward~\cite{grpo} and Language Rewards~\cite{deepseek2025}.}
  \label{fig:reward_frame}
  \vskip -0.1in
\end{figure}

\paragraph{Dual-Reward Design}
To mitigate the impact of noisy evaluation scores, we equip the selector with a dual-reward mechanism that restructures supervision signals instead of propagating raw values. As shown in Figure \ref{fig:reward_frame}, the mechanism combines a Preference-based Outcome Reward (POR) and a Context-Perception Reward (CPR); two lightweight auxiliary rewards enforce output format and language consistency similar as in~\cite{grpo}.

\paragraph{Preference-based Outcome Reward (POR)}
Following the analysis in Section \ref{sec:challenges_in_noisy_data_for_long_sequence_decision-making}, the rollout values produced by the MCTS evaluator are inevitably noisy: similar heuristics may receive widely different scores across runs, while heuristics with genuinely different performance can end up with almost identical scores. Such distortions mislead the learner and weaken the training signal.

\begin{figure}[!htbp]
  \centering
  \includegraphics[width=0.8\textwidth]{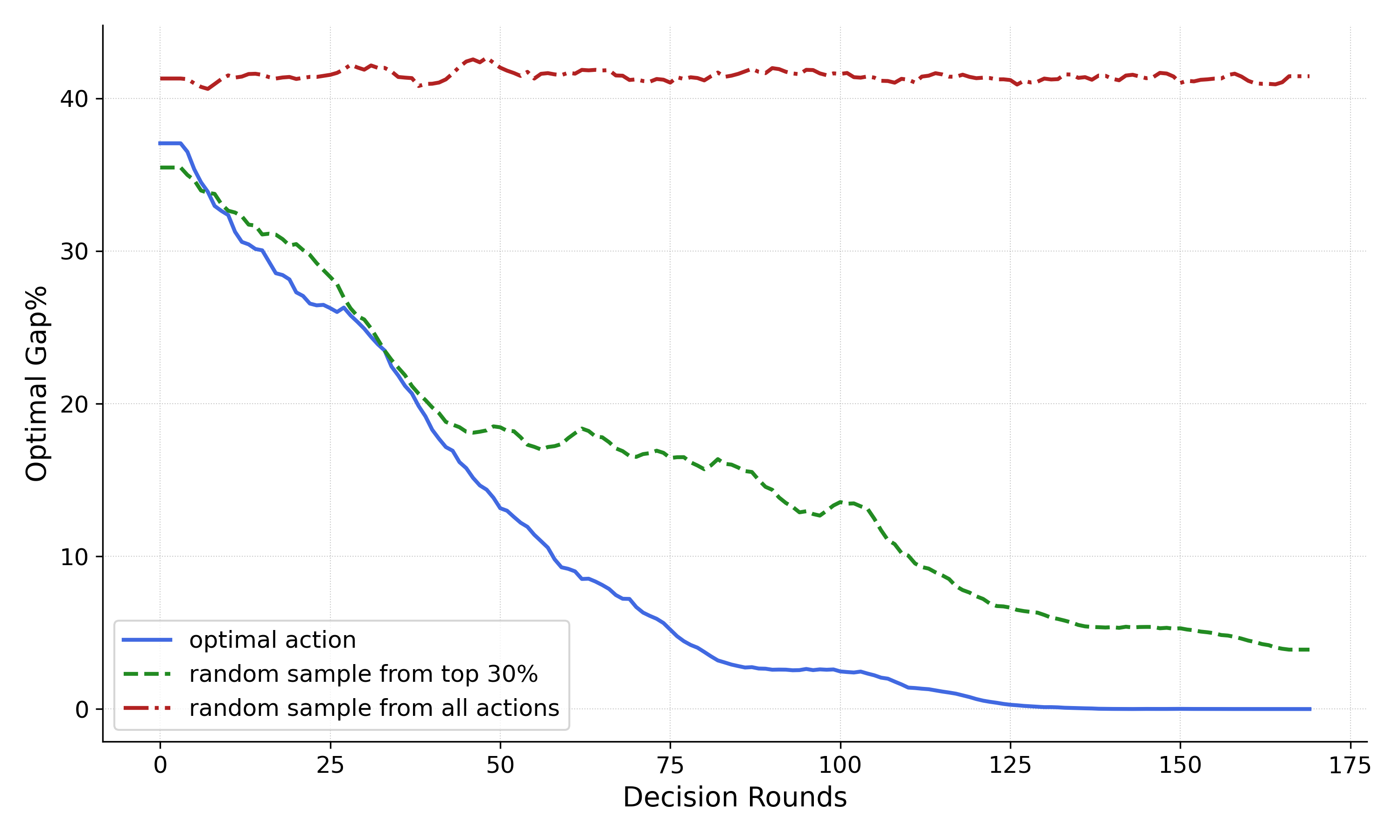}
  \caption{Effect of noisy rollout data on heuristic selection (\texttt{rd100} in TSPLIB~\cite{reinelt1991tsplib}). 
  Y-axis: expected optimality gap (lower is better) after completing the tour by random sampling. 
  X-axis: decision rounds. 
  Blue: always selecting the best heuristic (oracle). 
  Green: uniformly selecting from the top 30\% heuristics (positive set). 
  Red: random selection. 
  Selecting from the positive set almost matches the oracle and clearly outperforms random choice.}
  \label{fig:exp_reward}
\end{figure}

Figure \ref{fig:exp_reward} shows that, for our selection tasks, choosing any heuristic from a small positive set is nearly as good as always picking the single best one, whereas selecting from the complementary negative set quickly harms solution quality. In practice, mild noise is sufficient to interchange the ranks of neighbouring heuristics, yet it rarely pushes a truly positive heuristic into the negative set (or vice-versa) because the gap between the two sets is usually large.

These observations motivate a reward design that compresses score differences within the positive or negative group while enlarging the margin between the two groups. Let the rollout scores \(\{Q_H\}_{H\in\mathcal {H}}\) be sorted in descending order for the current state \(z\). Suppose the model proposes a heuristic \(\hat H\) ranked at position \(\ell\). Two thresholds \(n_{\text{pos}}<n_{\text{neg}}\le n=|\mathcal {H}|\) divide the list into a positive region, a negative region, and an error region. The preference-based outcome reward is then defined as
\begin{subequations}
\label{eq:por_reward}
\begin{align}
\ell &= \operatorname{rank}\bigl(z,\hat H,\{Q_H\}_{H\in\mathcal {H}}\bigr), \\[4pt]
R_{\mathrm{POR}}(z,\hat H) &=
\begin{cases}
  R_{p}\!\left(1-\dfrac{\ell-1}{n_{\text{pos}}}\right), & 1\le\ell\le n_{\text{pos}}, \\[6pt]
 -R_{n}\!\left(\dfrac{\ell-n_{\text{pos}}}{n_{\text{neg}}-n_{\text{pos}}}\right), & n_{\text{pos}}<\ell\le n_{\text{neg}}, \\[6pt]
 -R_{L}, & n_{\text{neg}}<\ell\le n .
\end{cases}
\end{align}
\end{subequations}

Linear interpolation reduces gaps inside each region, whereas the coefficients \(R_{p}\) and \(R_{n}\) amplify the discontinuity at the region boundary; heuristics in the error region receive a fixed penalty \(-R_{L}\). Consequently, POR is insensitive to small permutations inside the positive set but still yields a strong gradient whenever a heuristic crosses the positive/negative border.

The robustness of POR is demonstrated with a four-heuristic toy example (Table \ref{tab:toy_por}).  Heuristics are divided into a positive set (top two ranks) and a negative set (bottom two).  We compare POR ($R_{p}=R_{n}=1$) with the Normalized Rank Reward (NRR), a common RLHF baseline that linearly maps ranks to $[-1,1]$ \cite{ziegler2019fine,conti2018improving}. Under noise free scores, both rewards differentiate the two sets. When two middle scores are noised, NRR almost eliminates the gap between $\mathrm{H}_2$ and $\mathrm{H}_3$ ($0.2$ vs $-0.2$), while POR still keeps a clear margin ($0.5$ vs $-0.5$). Hence POR supplies a much steadier learning signal in the presence of measurement noise.

\begin{table}[h]
\centering
\caption{Toy example comparing POR with Normalized Rank Reward (NRR).  NRR maps ranks to $[-1,1]$. “Noisy” columns show the effect after perturbing the two middle scores.}
\label{tab:toy_por}
\small
\begin{tabular}{c|ccc|ccc}
\toprule
Heuristic & Actual score & Actual NRR & Actual POR & Noisy score & Noisy NRR & Noisy POR \\
\midrule
H\(_1\) & 1.0 &  1.0 & 1.0 & 1.0 &  1.0 &  1.0 \\
H\(_2\) & 0.8 &  0.6 & 0.5 & 0.6 &  0.2 &  0.5 \\
H\(_3\) & 0.2 & -0.6 & -0.5 & 0.4 & -0.2 & -0.5 \\
H\(_4\) & 0.0 & -1.0 & -1.0 & 0.0 & -1.0 & -1.0 \\
\bottomrule
\end{tabular}
\end{table}

\paragraph{Context-Perception Reward (CPR)}
CPR explicitly rewards the selector for understanding the environment, rather than for matching a potentially noisy scalar outcome. Recent evidence shows that even when the final reward signal is noisy or even random, models can still improve provided that their reasoning traces remain correct and receive feedback~\cite{shao2025spurious}. Moreover, empirical studies demonstrate that an accurate internal perception of the task state is often a stronger predictor of downstream performance than the magnitude of the terminal reward itself~\cite{khalifa2025process}. Motivated by these findings, CPR encourages the model to align its latent representation with the ground-truth state.

\begin{figure}[!htbp]
  \centering
  \includegraphics[width=\textwidth]{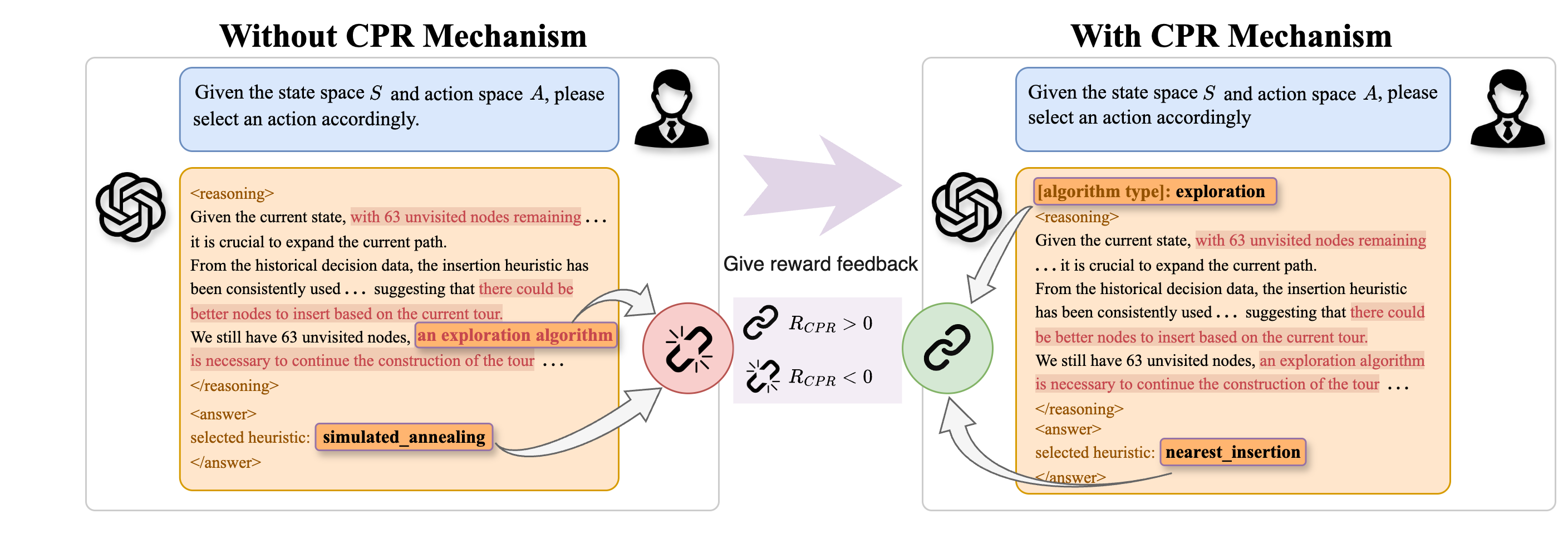}
  \caption{Impact of the CPR mechanism in rectifying qualitative judgment errors during LLM inference. The illustration shows how CPR guides the model toward accurate contextual assessments, mitigating errors that might otherwise remain uncorrected.}
  \label{fig:cpr}
\end{figure}

Formally, let $z=(z_1,\dots,z_m)$ be the true feature vector of the current problem state and $\hat z=(\hat z_1,\dots,\hat z_m)$ the selector’s own prediction of these $m$ features, we get:
\begin{equation}
R_{\mathrm{CPR}}(z,\hat z)=
\sum_{i=1}^{m}
\Bigl[
\mathbb{I}\!\bigl(\hat z_i = z_i\bigr)\,R_i^{+}
-\mathbb{I}\!\bigl(\hat z_i \neq z_i\bigr)\,R_i^{-}
\Bigr],
\label{eq:cpr_reward}
\end{equation}
where $\mathbb{I}(\cdot)$ is the indicator function and $R_i^{+}\!>\!0$ ($R_i^{-}\!>\!0$) denotes the positive (negative) reward tied to the $i$-th feature. Correctly perceived dimensions therefore yield positive feedback, while misperceptions are penalized, pushing the selector toward a faithful contextual understanding before it commits to a heuristic decision. Figure~\ref{fig:cpr} visualizes how CPR rectifies qualitative judgment errors and guides the model toward more accurate state assessments.

\paragraph{Training Procedure}
Under the dual-reward design, the selection model's parameters \(\theta\) are fine-tuned offline using the Group Relative Policy Optimization (GRPO)~\cite{grpo} method. GRPO's key approach involves leveraging inter-response preference relations to iteratively update the policy parameters \(\theta\). This framework aims to optimize the likelihood that the model generates response sequences leading to higher cumulative evaluation scores \(R_{\text{total}}\). The goal of this training process is to enable the selection model to proficiently generate heuristic sequences that yield superior performance scores, thus improving its effectiveness in real-world applications of combinatorial optimization. The Algorithm ~\ref{alg:grpo_heuristic} outlines the training process.

\begin{algorithm}[t]
  \small
  \caption{Fine-tuning Pipeline for Heuristic Selection}
  \textbf{Input} offline dataset $D_{\text{offline}}$; initial policy $\pi_{\theta_0}$;  
  reward weights $\lambda_{\text{POR}}, \lambda_{\text{CPR}}$ \textit{(optional $\lambda_{\text{base}}$)};  
  hyper-parameters $(G,\, M_{\text{epochs}},\, B_{\text{size}},\, \epsilon,\, \beta,\, \mu)$
  \begin{algorithmic}[1]
    \State $\pi_\theta \leftarrow \pi_{\theta_0}$ \Comment{Base selector policy}
    \For{epoch $e=1$ \textbf{to} $M_{\text{epochs}}$}
        \State Sample mini-batch $\{z_k\}_{k=1}^{B_{\text{size}}} \subset D_{\text{offline}}$
        \For{each state $z_k$}
            \State Sample $G$ heuristic sequences $\bigl\{H_k^{(g)}\bigr\}_{g=1}^{G} \sim \pi_\theta(\,\cdot \mid z_k)$
            \For{$g=1$ \textbf{to} $G$}
                \State Compute $R_{\text{POR}}(z_k, H_k^{(g)})$ \Comment{Refer equation~\eqref{eq:por_reward}}
                \State Compute $R_{\text{CPR}}(z_k)$ \Comment{Refer equation~\eqref{eq:cpr_reward}}
                \State Compute $R_{\text{base}}(z_k, H_k^{(g)})$ \Comment{Following the GRPO methodology}
                \State $R_{k}^{(g)} \leftarrow
                       \lambda_{\text{POR}} R_{\text{POR}} +
                       \lambda_{\text{CPR}} R_{\text{CPR}} +
                       \lambda_{\text{base}} R_{\text{base}}$
            \EndFor
            \State Estimate group-relative advantages $\widehat{A}_{k}^{(g)}=\;R_{k}^{(g)}-\frac{1}{G}\sum_{g'}R_{k}^{(g')}$ for all $g$. \hfill\Comment{GRPO core}
            \For{inner step $u=1$ \textbf{to} $\mu$}
                \State Update $\pi_\theta$ by maximizing GRPO objective with clip $\epsilon$ and KL penalty $\beta$
            \EndFor
        \EndFor
    \EndFor
  \end{algorithmic}
  \textbf{Output} fine-tuned parameters $\theta$
  \label{alg:grpo_heuristic}
\end{algorithm}

\section{Experiments}
\label{sec:experiments}

In this section, we conduct a comprehensive evaluation of HeurAgenix under diverse settings to assess its performance, robustness, and adaptability. We explore heuristic evolution outcomes in Section~\ref{sec:evolution_experiments}, analyze problem solving efficiency using LLMs in Section~\ref{sec:problem_solving_with_llm}, and investigate the capabilities of fine-tuned models in Section~\ref{sec:problem_solving_with_fine_tuned_model}. 
To ensure fairness and consistency throughout our experiments, we adhere to the following setup and more detailed parameter settings can be found in Appendix~\ref{sec:detailed_parameter_settings}:
\begin{enumerate}
    \item \textbf{Foundation Model:} In the heuristic evolution phase, GPT-4o (version: 2024-11-20) is employed as the foundation model for heuristic evolution and problem solving phase. To ensure fair comparisons, we consistently use this model across all LLM-based hyper-heuristics, including EoH and ReEvo. In the problem solving phase using LLMs, GPT-4o continues as the primary selector, while Qwen-7B~\cite{bai2023qwen} serves as the base model for fine-tuning in the fine-tuned model experiments.

    \item \textbf{API Calls:} During heuristic evolution, all hyper-heuristics evolve until they reach 2,000 API calls, ensuring consistency across methods. In problem solving, our method utilizes \(\lceil \frac{N}{M}\rceil + 2\) API calls—where \(N\) is max heuristic steps and related to problem size and \(M\) is fixed at 5, with 2 additional calls for problem description and heurisitic pool introduction.

    \item \textbf{Execution Time:} Each test instance is allocated a maximum runtime of two hours to exploit the advantages of search-based methods.

    \item \textbf{Metric:} Solution performance is quantified by the optimality-gap metric, \(\text{gap} = \frac{v - v^{u}}{v^{u}} \times 100\%\), where \(v\) is the obtained solution value and \(v^{u}\) is the known optimal (or best-known) value. To reduce variance, every experiment is run three times.

    \item \textbf{Platform:} Experiments are conducted on a GNU/Linux system with a 5.15.0-121-generic kernel, Intel(R) Xeon(R) processor, NVIDIA RTX A6000 (48G) GPU, and CUDA 12.2.

\end{enumerate}
We evaluate our framework across the following combinatorial optimization problems:

\begin{enumerate}
    \item \textbf{Traveling Salesman Problem (TSP):} Utilizes sub-datasets from TSPLIB~\cite{reinelt1991tsplib} to identify the shortest route visiting each city once.
    
    \item \textbf{Capacitated Vehicle Routing Problem (CVRP):} Employs the largest instances from the first six series of CVRPLIB~\cite{CVRPLIB} to optimize delivery routes with capacity constraints.
    
    \item \textbf{Multiple Knapsack Problem (MKP):} Uses instances from mknapcb1 and mknapcb4 in the OR-Library~\cite{beasley1990or} to maximize item value across multiple knapsacks.
    
    \item \textbf{Job Shop Scheduling Problem (JSSP):} Leverages the first 20 instances from the OR-Library~\cite{beasley1990or} to minimize total processing time for jobs across machines.
    
    \item \textbf{MaxCut Problem (MaxCut):} Uses the first 10 instances from Optsicom~\cite{corberan2006optsicom} to partition graph vertices and maximize the edge weight sum between vertex sets.
\end{enumerate}

\subsection{Evolution Experiments}
\label{sec:evolution_experiments}

In this section, we evaluate the effectiveness of our heuristic evolution strategy on five combinatorial optimization problems. Since the evolution interfaces of ReEvo and EoH are limited to the nearest neighbor heuristic in TSP, we compare with these two baselines only in that setting; for all other problems, we contrast the original seed heuristics with the evolved variants produced by HeurAgenix.
\begin{figure}[H]
  \centering
  \includegraphics[width=0.8\textwidth]{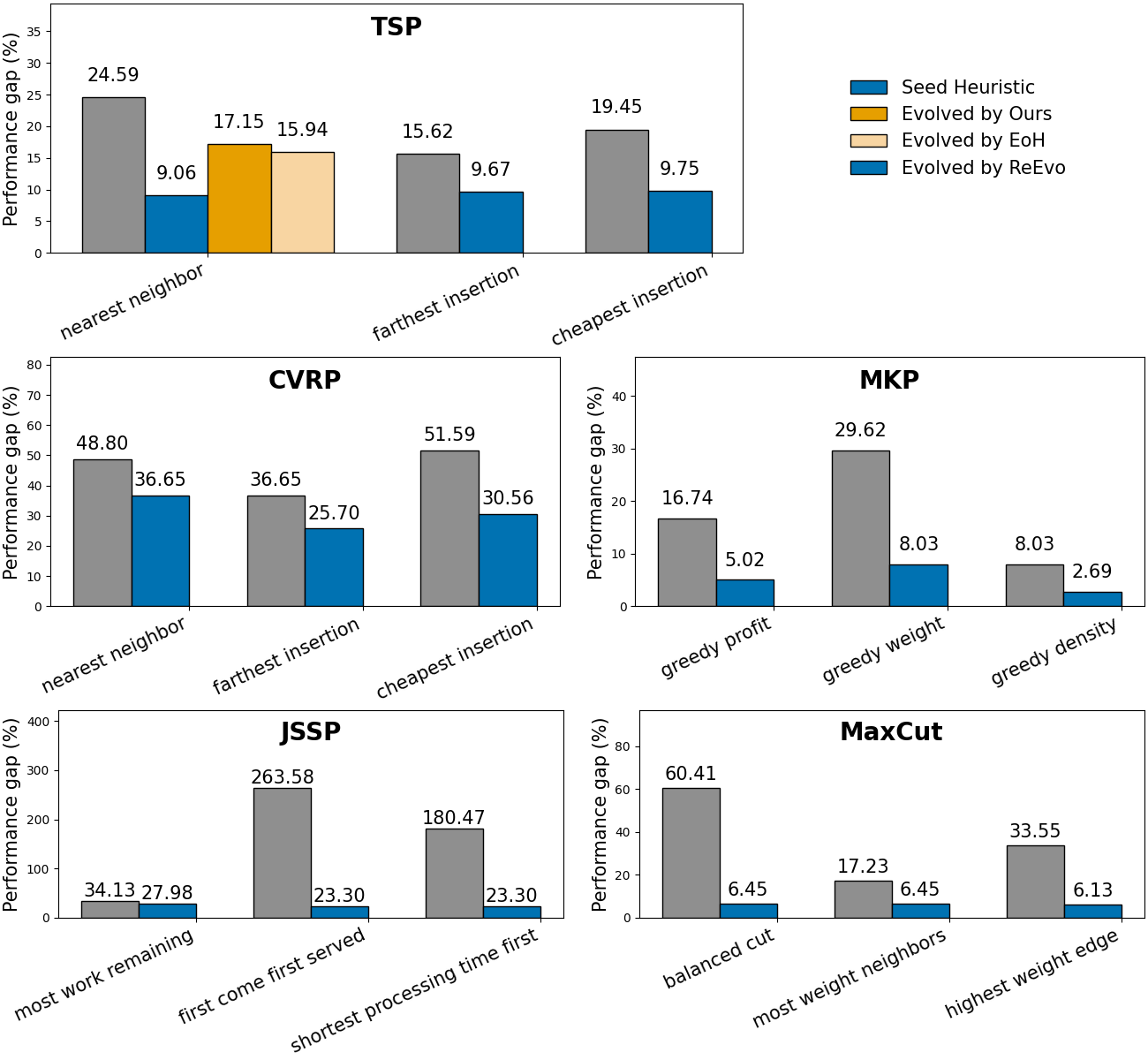}
  \caption{Average optimality gaps (\%; lower is better) before and after heuristic evolution on five representative CO problems. Complete results for all instances appear in Appendix~\ref{sec:detailed_evolution_results}.}
  \label{fig:evolution_overview}
\end{figure}

As figure \ref{fig:evolution_overview} the shows, HeurAgenix consistently and substantially reduces the gap, and even very basic seed heuristics can be transformed into highly competitive solvers through our automatic evolution pipeline. Detailed evolution results are provided in Table~\ref{tab:evolution_result} in Appendix~\ref{sec:detailed_evolution_results}.

\subsection{Problem Solving with LLM}
\label{sec:problem_solving_with_llm}
We benchmark HeurAgenix on five standard CO benchmarks, always using GPT-4o as the heuristic selector for the evolved heuristic pool, and contrast its performance with both traditional methods and the strongest available LLM-based hyper-heuristics.

For TSP we compare against Guided Local Search (GLS)~\cite{voudouris2010guided}, Ant Colony Optimization (ACO)~\cite{dorigo2007ant}, and OR-Tools~\cite{google2020ortools}, as well as three language-based hyper-heuristics: EoH~\cite{liu2024evolution}+GLS and ReEvo~\cite{ye2024reevo}+GLS, which keep the GLS framework but let EoH or ReEvo evolve its penalty function, and ReEvo+ACO, which uses ReEvo to refine the pheromone update rules of ACO. For CVRP we report ACO, OR-Tools, and ReEvo+ACO. For MKP we include ACO, ReEvo+ACO, the quantum-inspired algorithm QICSA~\cite{layeb2011novel}, and Particle Swarm Optimization (PSO)~\cite{eberhart1995particle}, with the QICSA and PSO numbers copied from Huang et al.~\cite{Huang2022ICIC}. The JSSP baselines are ACO, PSO, and the Grey Wolf Optimizer (GWO)~\cite{mirjalili2014grey} as reported by van Hoorn et al.~\cite{van2016dynamic}. The MaxCut baselines are Scatter Search (SS)~\cite{marti2009advanced}, CirCut~\cite{burer2002rank}, and VNSPR~\cite{festa2002randomized}, which are taken from Myklebust et al.~\cite{myklebust2015solving}.

\begin{figure}[H]
  \centering
  \includegraphics[width=0.8\textwidth]{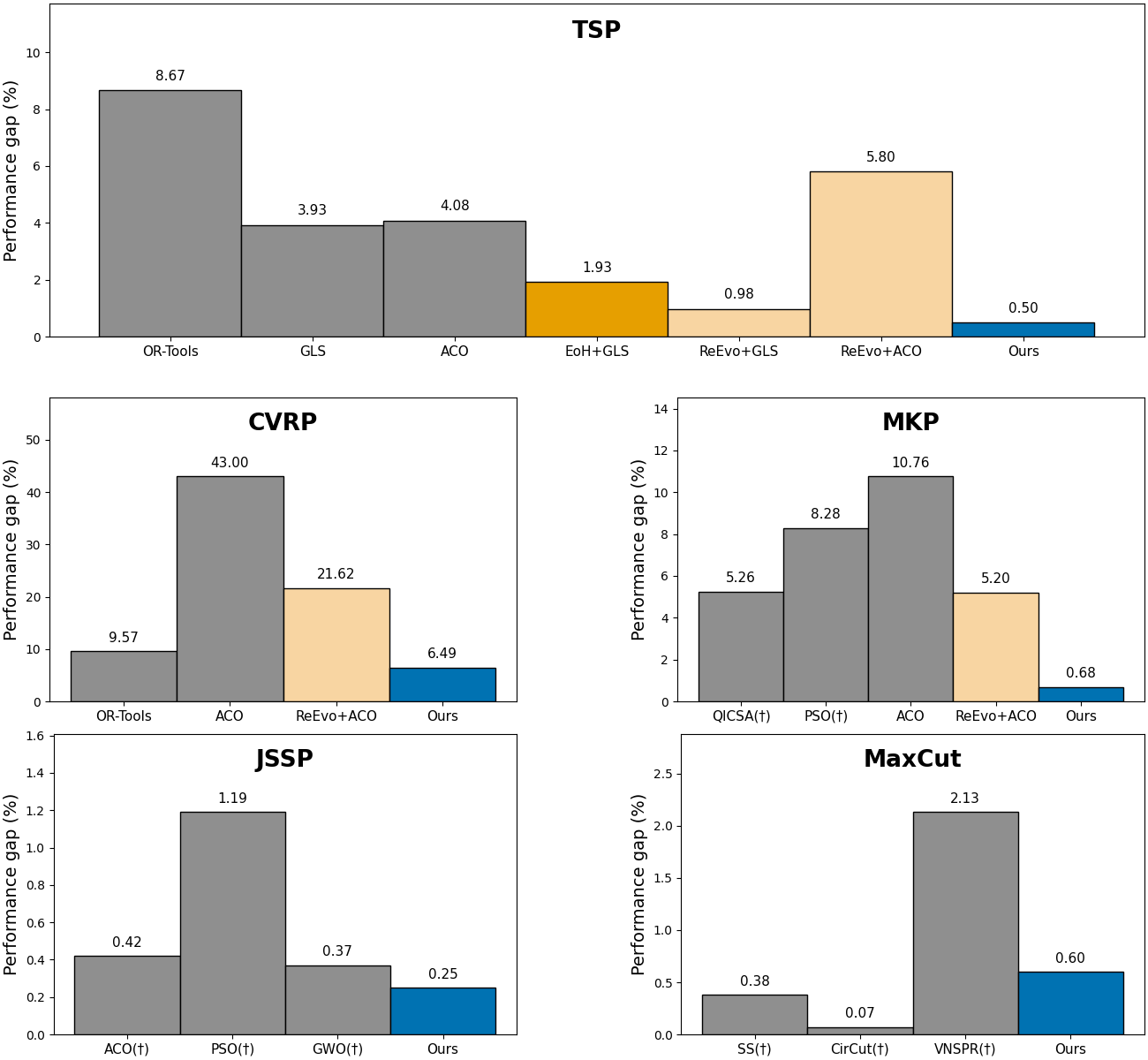}
  \caption{Average optimality gap (\%; lower is better) of HeurAgenix and baselines on five CO benchmarks. Dark-blue bars correspond to HeurAgenix (Ours), draw-orange bars to EoH related
  method, light-orange bars to ReEvo related methods, and gray bars to traditional methods.  
A dagger (†) after a method name indicates that the result is copied from the original publication.  
Complete results for all instances appear in Appendix~\ref{sec:detailed_problem_solving_results_with_llm}.}
  \label{fig:problem_solving_with_llm_overview}
\end{figure}

As shown in Figure~\ref{fig:problem_solving_with_llm_overview}, HeurAgenix decisively outperforms the existing language-based hyper-heuristics and matches or surpasses the specialized methods. Detailed performance results are provided in Table~\ref{tab:result_with_llm} in Appendix~\ref{sec:detailed_problem_solving_results_with_llm}.

\subsection{Problem Solving with a Fine-tuned Model}
\label{sec:problem_solving_with_fine_tuned_model}

We next investigate whether a lightweight model, fine-tuned under our dual-reward scheme, can serve as an effective online selector.  All experiments are conducted on TSPLIB instances, where the selector must choose from the identical evolved heuristic pool.

\begin{table}[H]
    \vskip -0.1in
    \begin{minipage}[t]{0.53\textwidth}
        \vspace{0pt}
        \centering
        \small
        \caption{Per-instance optimality gaps on TSPLIB (\%; lower is better) when the selector is a mainstream LLM. \textbf{Bold} marks the best result. Variance is omitted when it equals~0.}
        \label{tab:comparison_with_llm}
        \resizebox{\columnwidth}{!}{
          \begin{tabular}{l|cccc}
            \toprule
            Instance & GPT-4o & OpenAI O3 & DeepSeek-R1 & Ours \\
            \midrule
            kroA100 & \textbf{0} & \textbf{0} & \textbf{0} & \textbf{0} \\
            kroA150 & \textbf{0} & \textbf{0} & \textbf{0} & \textbf{0} \\
            kroB100 & \textbf{0} & \textbf{0} & \textbf{0} & \textbf{0} \\
            kroB200 & 0.11 & 0.10 $\pm$ 0.1 & \textbf{0} & 0.30 $\pm$ 0.1 \\
            kroC100 & \textbf{0} & \textbf{0} & \textbf{0} & \textbf{0} \\
            bier127 & 1.29 $\pm$ 0.6 & \textbf{0.22} $\pm$ 0.1 & 1.01 $\pm$ 0.1 & 1.09 $\pm$ 0.2 \\
            tsp225 & 0.23 $\pm$ 0.1 & 0.29 $\pm$ 0.1 & \textbf{0.20} & 0.24 $\pm$ 0.2 \\
            a280 & 0.16 $\pm$ 0.1 & \textbf{0} & 0.10 $\pm$ 0.1 & 0.91 $\pm$ 0.4 \\
            pcb442 & 2.10 $\pm$ 0.5 & \textbf{0.59} $\pm$ 0.1 & 0.80 $\pm$ 0.5 & 0.79 $\pm$ 0.2 \\
            gr666 & \textbf{0.93} $\pm$ 0.2 & 1.50 $\pm$ 0.3 & 1.19 $\pm$ 1.3 & 1.13 $\pm$ 0.9 \\
            pr152 & 0.23 $\pm$ 0.2 & \textbf{0.13} & 0.16 & 0.19 $\pm$ 0.1\\
            pr1002 & 1.78 $\pm$ 0.6 & 1.33 $\pm$ 0.4 & 1.30 $\pm$ 0.3 & \textbf{1.00} $\pm$ 0.3 \\
            pr2392 & 1.08 $\pm$ 0.5 & \textbf{0.87} $\pm$ 0.3 & 1.09 $\pm$ 0.7 & 0.92 $\pm$ 0.3 \\
            \midrule
            Average gap & 0.61 & \textbf{0.39} & 0.45 & 0.50 \\
            \bottomrule
          \end{tabular}}
    \end{minipage}
    \hfill
    \begin{minipage}[t]{0.43\textwidth}
        \vspace{0pt}
        \centering
        \small
        \caption{Ablation studies based on Qwen-7B. Optimality gaps on TSPLIB (\%; lower is better).  \textbf{Bold} marks the best per row. Variance is omitted when it equals~0.}
        \label{tab:ablation_studies}
        \resizebox{\columnwidth}{!}{
            \begin{tabular}{l|ccc}
            \toprule
            Instance & raw & GRPO & Ours \\
            \midrule
            kroA100 & \textbf{0} & 0.21 $\pm$ 0.1 & \textbf{0} \\
            kroA150 & 0.80 $\pm$ 0.1 & 0.91 $\pm$ 0.1 & \textbf{0} \\
            kroB100 & \textbf{0} & \textbf{0} & \textbf{0} \\
            kroB200 & 2.82 $\pm$ 1.4 & 2.02 $\pm$ 1.1 & \textbf{0.30} $\pm$ 0.1 \\
            kroC100 & 0.44 $\pm$ 0.1 & 0.10 & \textbf{0} \\
            bier127 & 2.82 $\pm$ 1.4 & 3.01 $\pm$ 1.1 & \textbf{1.09} $\pm$ 0.2 \\
            tsp225 & 5.22 $\pm$ 3.4 & 4.01 $\pm$ 2.4 & \textbf{0.24} $\pm$ 0.2 \\
            a280 & 4.82 $\pm$ 1.4 & 3.31 $\pm$ 0.4 & \textbf{0.91} $\pm$ 0.4 \\
            pcb442 & 3.00 $\pm$ 1.9 & 2.21 $\pm$ 1.9 & \textbf{0.79} $\pm$ 0.2 \\
            gr666 & 8.02 $\pm$ 5.4 & 4.82 $\pm$ 2.2 & \textbf{1.13} $\pm$ 0.9 \\
            pr152 & 1.49 $\pm$ 0.4 & 1.12 $\pm$ 0.3 & \textbf{0.19}$\pm$ 0.1 \\
            pr1002 & 9.98 $\pm$ 8.4 & 5.02 $\pm$ 2.7 & \textbf{1.00} $\pm$ 0.3 \\
            pr2392 & 10.86 $\pm$ 7.4 & 8.21 $\pm$ 7.3 & \textbf{0.92} $\pm$ 0.3 \\
            \midrule
            Average gap & 5.01 & 4.39 & \textbf{0.59} \\
            \bottomrule
          \end{tabular}}
    \end{minipage}
    \vskip -0.1in
\end{table}

\paragraph{Comparison to mainstream proprietary LLMs.}
Table~\ref{tab:comparison_with_llm} contrasts our fine-tuned \textbf{Qwen-7B} with three popular closed-source models operating in zero-shot mode (GPT-4o, OpenAI O3, DeepSeek-R1).  Despite its far smaller size and cost, the fine-tuned model achieves accuracy on par with the strongest proprietary alternatives.

\paragraph{Ablation studies.}
Table~\ref{tab:ablation_studies} studies the contribution of our dual-reward fine-tuning.  Starting from the raw Qwen-7B, we compare (a) vanilla GRPO and (b) our POR+CPR rewards.  The dual-reward variant reduces the average gap from 5.01\% to 0.59\%, and dominates GRPO on every instance.

\begin{figure}[H]
  \centering
  \includegraphics[width=0.7\linewidth]{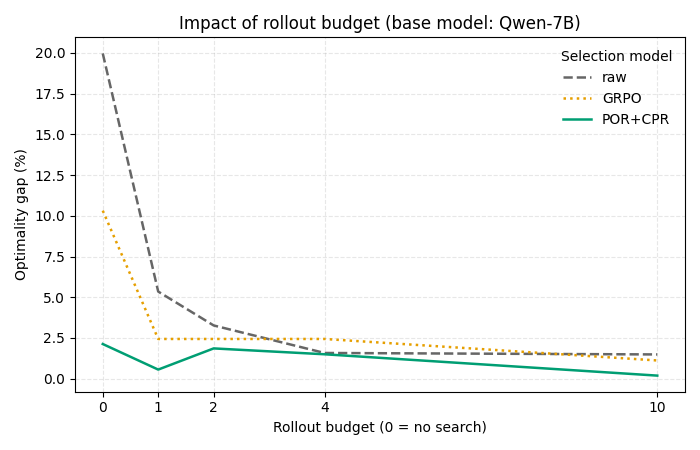}
  \caption{Impact of rollout budget on pr152 from TSPLIB~\cite{reinelt1991tsplib}.
  X-axis: Monte Carlo samples per candidate heuristic (\textbf{rollout budget}; 0 = direct use of the LLM-proposed heuristic with no search). Base model is Qwen-7B.
  Y-axis: optimality gap (\%; Lower is better).}
  \label{fig:tts_influence}
\end{figure}

\paragraph{Effect of test-time search.}
We next quantify the benefit of test-time search. During inference, the selector may perform a small Monte Carlo search: for each candidate heuristic proposed by the LLM, it rolls out $k$ random completions and takes the average score.  
We refer to $k$ as the \textbf{rollout budget}. Figure~\ref{fig:tts_influence} plots the optimality gap versus the rollout budget on the pr152 instance. Larger budgets consistently reduce the gap as expected, and our dual-reward fine-tuned selector (POR\,+\,CPR) dominates all baselines at any fixed budget. 

\section{Conclusion}
\label{sec:conclusion}
We have introduced \textbf{HeurAgenix}, an end-to-end, LLM-driven hyper-heuristic framework that automatically evolves diverse heuristics and selects among them online. A contrastive, data-driven evolution phase discovers reusable improvement patterns without external solvers, while an adaptive selector—implemented either with an off-the-shelf LLM or a fine-tuned, lightweight model—leverages test-time scaling and a dual-reward mechanism to remain robust under noisy supervision. On standard CO benchmarks, HeurAgenix consistently outperforms existing LLM-based hyper-heuristics and matches or exceeds specialized solvers, demonstrating strong scalability and domain generality.

Several avenues remain for future work. First, our empirical study has so far been limited to the Qwen-7B backbone, and we will evaluate HeurAgenix across a broader spectrum of model sizes and architectures to assess robustness and cost–performance trade-offs. Then, the current POR relies on a manually chosen positive/negative split, and we plan to devise adaptive, data-driven schemes that adjust these thresholds dynamically during training. Last, while the selector now targets heuristic choice for classical CO tasks, its principle—sequentially picking an operator in a finite operation space with only terminal rewards—naturally extends to more general finite-state Markov Decision Processes(MDP), and exploring this wider class of decision problems is an important next step.

\clearpage
\bibliographystyle{plain}
{\small
  \bibliography{references}
}
\newpage
\appendix

\section{Heuristic Format}
\label{sec:heuristic_format}

In order to facilitate dynamic heuristic switching, all heuristics within the heuristic pool adhere to a uniform format. Each heuristic is implemented as a Python function, characterized by the following structure: 

\captionsetup[lstlisting]{labelformat=empty}
\lstset{
  language=Python,
  basicstyle=\ttfamily\small,
  keywordstyle=\color{blue},
  commentstyle=\color{gray},
  stringstyle=\color{black},
  showstringspaces=false,
  breaklines=true,
  frame=single,
  escapeinside={(*@}{@*)}
}

\begin{lstlisting}[language=Python, caption={Nearest Neighbor Heuristic}]
def nearest_neighbor_f91d(problem_state: dict, algorithm_data: dict, **kwargs) -> Tuple[Operator, dict]:
    """...
    Args:
        ...
    Returns:
        Tuple[Operator, dict]: ...
    """
    # Retrieve necessary data from problem_state
    ...
    # If the current solution is empty, start from the first unvisited node.
    if not current_solution.tour:
        start_node = unvisited_nodes[0]
        return AppendOperator(node=start_node), {}

    # Find the nearest neighbor to the last visited node.
    nearest_node = None
    min_distance = float('inf')
    for node in unvisited_nodes:
        distance = distance_matrix[last_visited][node]
        if distance < min_distance:
            nearest_node = node
            min_distance = distance
    position=len(current_solution.tour)
    return InsertOperator(node=nearest_node, position=position), {}
\end{lstlisting}

Some additional remarks:
1. The function name ends with a unique 4-digit identifier (\texttt{f91d} in this example) to avoid naming conflicts. \\
2. The input consists of \texttt{problem\_state}, and \texttt{algorithm\_data}, which store problem state, and control parameters, respectively. \\
The output consists of the current solution's operation and additional information. In this example, \texttt{AppendOperator(node)} adds a node to the end of the current tour and no more information need be passed. Other TSP heuristics may use \texttt{InsertOperator}, \texttt{SwapOperator}, \texttt{ReverseSegmentOperator}, etc.

\section{Initial Heuristic Generation}
\label{sec:initial_heuristic_generation}

For CO problems, the evolutionary process requires initialial seed heuristic algorithms. These can be manually crafted or be generated by HeurAgenix. 
For classic heuristics, HeurAgenix can \textbf{generate from LLM}, where heuristics are produced using the internal knowledge of language models.
HeurAgenix can also \textbf{learn from literature}, which involves reading abstracts to determine relevance, selecting interesting sections, and then generating heuristics based on the information extracted from relevant sections.
When dealing with new problems, HeurAgenix can \textbf{transfer heuristics from known problems}. The detailed steps for transferring heuristics from related problems are as follows:

\textbf{Decompose new and known problems}: The LLM decomposes both the new problem and source problems into their respective components. \\
\textbf{Match components}: The LLM compares the components of the new problem with those of known problems to determine if heuristics from these problems can be leveraged. \\
\textbf{Analyze heuristics from known problems}: If applicable, the LLM reads the heuristics from these known problems. \\
\textbf{Evaluate and transfer}: For each heuristic, if the LLM determines it is transferable, it translates the components and begins the transfer process; otherwise, it skips this heuristic. \\

\section{Heuristic Evolution Example}
\label{sec:heuristic_evolution_example}

\captionsetup[lstlisting]{labelformat=empty}
\definecolor{codegreen}{rgb}{0,0.6,0}
\definecolor{codered}{rgb}{0.6,0,0}
\lstset{
  basicstyle=\footnotesize\ttfamily\color{black}, 
  numberstyle=\tiny\color{gray}, 
  stepnumber=1, 
  numbersep=5pt, 
  xleftmargin=2em, 
  frame=tb, 
  columns=fullflexible,
  keepspaces=true,
  showstringspaces=false, 
  showtabs=false, 
  tabsize=2, 
  keywordstyle=\color{black},      
  commentstyle=\color{black},      
  stringstyle=\color{black},       
  language=, 
  moredelim=**[is][\color{codered}]{@}{@}, 
  moredelim=**[is][\color{codegreen}]{^}{^} 
}

The following diff listings capture the full five-rounds evolution (see Fig.~\ref{fig:evolution_example}) of the nearest neighbor heuristic for the TSP. Lines prefixed with “–” (rendered in red) were removed, while lines prefixed with “+” (rendered in green) were introduced. \\

\begin{lstlisting}[caption={Evolution Round 1: Adjust Initial Node}]
if not current_solution.tour:
@-     start_node = unvisited_nodes[0]@
^+     avg_distances = [^
^+         np.mean([distance_matrix[i][j]^
^+                  for j in range(node_num) if i != j^
^+         ])^
^+         for i in range(node_num)^
^+     ]^
^+     sub_central_node = np.argsort(avg_distances)[1]^
^+     start_node = min(unvisited_nodes,^
^+         key=lambda node:^
^+             distance_matrix[sub_central_node][node])^
...
\end{lstlisting}

\begin{lstlisting}[caption={Evolution Round 2: Evaluate Node Selection}]
...
for node in unvisited_nodes:
@-     score = distance_matrix[last_visited][node]@
^+     future_cost = sum([^
^+         distance_matrix[node][unvisited] ^
^+             for unvisited in unvisited_nodes if unvisited != node^
^+     ])^
^+     immediate_cost = distance_matrix[last_visited][node]^
^+     score = immediate_cost + ^
^+         problem_state["visited_num"] / node_num / node_num * future_cost^
    if distance < best_score:
        ...
\end{lstlisting}

\begin{lstlisting}[caption={Evolution Round 3: Adjust Insert Location}]
...
for node in unvisited_nodes:
    future_cost = sum([
        distance_matrix[node][unvisited] 
        for unvisited in unvisited_nodes 
        if unvisited != node
    ])
@-     score = immediate_cost + @
@-         problem_state["visited_num"] / node_num / node_num * future_cost@
^+     for i in range(len(current_solution.tour) + 1):^
^+         prev_node = current_solution.tour[i - 1]^
^+         next_node = current_solution.tour[i]^
^+         immediate_cost = distance_matrix[prev_node][node] + ^
^+             distance_matrix[node][next_node] - ^
^+             distance_matrix[prev_node][next_node]^
^+         position = i^
...
\end{lstlisting}

\begin{lstlisting}[caption={Evolution Round 4: Limit Candidate Nodes}]
...
^+ threshold_factor = kwargs.get("threshold_factor", 0.70)^
^+ percentage_range = kwargs.get("percentage_range", 0.20)^
^+ # Calculate average distance from the last visited node to unvisited nodes^
^+ avg_distance = np.mean([^
^+     distance_matrix[last_visited][node] for node in unvisited_nodes^
^+ ])^

^+ # Find nearest unvisited node to the last visited node^
^+ nearest_node = min(unvisited_nodes, ^
^+    key=lambda node: distance_matrix[last_visited][node]^
^+    )^
^+ nearest = distance_matrix[last_visited][nearest_node]^

^+ # Prioritize inserting the nearest node if its distance is significantly shorter^
^+ if nearest < threshold_factor * avg_distance:^
^+     return InsertOperator(node=nearest_node, position=len(current_solution.tour))^

^+ # Evaluate nodes with comparable distances^
^+ comparable_nodes = [node ^
^+     for node in unvisited_nodes ^
^+         if distance_matrix[last_visited][node] <= (1 + percentage_range) * nearest^
^+ ]^

@- for node in unvisited_nodes:@
^+ for node in comparable_nodes:^
    ...
\end{lstlisting}

\begin{lstlisting}[caption={Evolution Round 5: Periodic Apply 2-opt}]
... 
^+ apply_2opt_frequency = kwargs.get("apply_2opt_frequency", 5)^
^+ N = len(current_solution.tour)^
^+ if len(current_solution.tour) > 2:^
^+    if len(current_solution.tour) % apply_2opt_frequency == 0:^
^+        best_delta = 0^
^+        best_pair = None^
^+        for i in range(N - 1):^
^+            for j in range(i + 2, N):^
^+                if j == N - 1 and i == 0:^
^+                    continue^

^+             a = current_solution.tour[i]^
^+             b = current_solution.tour[(i+1) % N]^
^+             c = current_solution.tour[j]^
^+             d = current_solution.tour[(j+1) % N]^
^+             current_cost = distance_matrix[a][b] + distance_matrix[c][d]^
^+             new_cost = distance_matrix[a][c] + distance_matrix[b][d]^
^+             delta = new_cost - current_cost^
 
^+             if delta < best_delta:^
^+                 best_delta = delta^
^+                 best_pair = (i + 1, j)^
 
^+     if best_pair:^
^+         return ReverseSegmentOperator([best_pair]), {}^
...
\end{lstlisting}

\section{Monte Carlo Evaluation Strategy}
\label{sec:monte_carlo_evaluation_strategy}

The Monte Carlo Evaluation Strategy aims to evaluate the effectiveness of heuristic algorithms by simulating their performance across multiple problem states. This strategy involves running a series of tests where each heuristic algorithm is assessed based on its ability to improve the solution quality from a given state. The process ensures that each heuristic is tested in a consistent environment, allowing for a fair comparison of their relative strengths and weaknesses.

As shown in Algorithm~\ref{alg:monte_carlo_evaluation_strategy}, the pseudocode outlines the detailed steps involved in this strategy:

\begin{algorithm}[H]
  \small
  \caption{Monte Carlo Evaluation Strategy}
  \textbf{Input} problem instance $d$ with initial state $z_0$;
  test heuristic $H_{\text{test}}$; heuristic pool $\mathcal{H}$;
  application frequency $M$; number of evaluations $T$ \\[2pt]
  \textbf{Output} estimated quality $Q_{H_{\text{test}}}$
  \begin{algorithmic}[1]
    \State Initialize list $\mathit{metrics}\leftarrow[\,]$
    \For t = 1 … T
        \State Apply $H_{\text{test}}$ exactly $M$ times to $z_0$ to obtain $z_1$
        \State $z' \leftarrow z_1$
        \While{the selected heuristic continues to improve the solution}
            \State Randomly choose $H \in \mathcal{H}$
            \State Apply $H$ once to update $z'$
        \EndWhile
        \State Append the final-solution metric to $\mathit{metrics}$
    \EndFor
    \State $Q_{H_{test}} \leftarrow \operatorname{mean}(\mathit{metrics})$
    \State \Return $Q_{H_{test}}$
  \end{algorithmic}
  \label{alg:monte_carlo_evaluation_strategy}
\end{algorithm}

\section{Detailed Parameter Settings}
\label{sec:detailed_parameter_settings}
This section introduces the detailed experiment settings including heuristic evolution and problem solving.
\begin{itemize}
    \small
    \item \textbf{Experiment platform}
    \begin{itemize}
        \item OS: GNU/Linux
        \item kernel: 5.15.0-121-generic
        \item Architecture: x86\_64
        \item Processor: Intel(R) Xeon(R)
        \item GPU: 4 × NVIDIA RTX A6000 (48G)
        \item CUDA: 12.2
    \end{itemize}
    \item \textbf{LLM parameters}
    \begin{itemize}
        \item Model: GPT-4o\_2024-11-20(2024-05-01-preview)
        \item Model: OpenAI-o3\_2025-04-16(2024-05-01-preview)
        \item Model: DeepSeek-R1
        \item Model: Qwen2.5-7B-Instruct-1M
        \item Temperature: 0.7
        \item Top-p: 0.95
        \item Max tokens: 1600
    \end{itemize}
    \item \textbf{Evolution parameters}
    \begin{itemize}
        \item Max API calls for each problem: 2000
        \item Train size for each problem: 20
        \item Max perturbation trials $P$: 1000
        \item Perturbation ratio (|K| / N): 0.1
        \item Max refinement iterations $T$: 5
    \end{itemize}
    \item \textbf{Solving parameters}
    \begin{itemize}
        \item Time limitation: 2 hours
        \item Selection frequency (M): 5
        \item Monte Carlo search times: 10
        \item Problem state context length: 1000
    \end{itemize}
    \item \textbf{Fine-tuning model parameters}
    \begin{itemize}
        \item LoRA Rank ($\alpha$): 32
        \item Optimizer: Paged AdamW (8-bit) (`paged\_adamw\_8bit`)
        \item Learning Rate: $1 \times 10^{-6}$
        \item Adam $\beta_1$: 0.9
        \item Adam $\beta_2$: 0.99
        \item Weight Decay: 0.1
        \item Learning Rate Scheduler: Cosine (`cosine`)
        \item Warmup Ratio: 0.1
        \item Max. Gradient Norm: 0.1
        \item Mixed Precision: BF16 (if supported by hardware, else FP16)
        \item Per-Device Training Batch Size: 1
        \item Gradient Accumulation Steps: 1
        \item Number of Training Epochs: 1
        \item Maximum Training Steps: -1 (disabled; training duration set by epochs)
        \item Inference Utility for Generation: vLLM (enabled via `use\_vllm=True`)
        \item Generations per Prompt ($N_G$): 12
        \item Max. Prompt Length: 2048 tokens
        \item Max. Completion Length: 768 tokens
    \end{itemize}
    \item \textbf{TSP}
    \begin{itemize}
        \item Data source: \url{http://comopt.ifi.uni-heidelberg.de/software/TSPLIB95/tsp/}
        \item Train instances: 20 generated instances
        \item Validation instances: brg180, eil101, gr202, pr124, pr152, rd100, u159
        \item Test instances: kroA100, kroA150, kroB100, kroB200, kroC100, bier127, tsp225, a280, pcb442, gr666, pr1002, pr2392
        \item Basic heuristics: cheapest insertion, farthest insertion, greedy algorithm, greedy randomized adaptive search procedure grasp, insertion heuristics, nearest insertion, nearest neighbor, random pairwise insertion, 2opt,  3opt
        \item Problem states: average\_distance, min\_distance, max\_distance, std\_dev\_distance, node\_num, current\_path\_length, remaining\_nodes, current\_cost, average\_edge\_cost, last\_edge\_cost, std\_dev\_edge\_cost, solution\_validity, min\_edge\_cost\_remaining, max\_edge\_cost\_remaining
        \item Operators: AppendOperator, InsertOperator, SwapOperator, ReverseSegmentOperator, RelocateOperator
    \end{itemize}
    \item \textbf{CVRP}
    \begin{itemize}
        \item Data source: \url{http://vrp.galgos.inf.puc-rio.br/index.php/en/}
        \item Train instances: 20 generated instances
        \item Validation instances: A-n63-k10, B-n67-k10, E-n76-k10, F-n45-k4, M-n101-k10, P-n70-k10, X-n101-k25
        \item Test instances: A-n80-k10, B-n78-k10, E-n101-k14, F-n135-k7, M-n200-k17, P-n101-k4
        \item Basic heuristics: farthest insertion, greedy, min cost insertion, nearest neighbor, node shift between routes, petal algorithm, saving algorithm, 2 opt, 3 opt
        \item Problem states: average\_demand, demand\_variance, upper\_triangle\_indices, upper\_triangle\_distances, average\_distance, max\_distance, min\_distance, distance\_variance, vehicle\_capacity\_utilization, node\_to\_vehicle\_ratio, average\_route\_length, max\_route\_length, min\_route\_length, std\_dev\_route\_length, average\_route\_cost, total\_demand\_served, average\_vehicle\_load, average\_remaining\_vehicle\_capacity, number\_of\_unvisited\_nodes, average\_unvisited\_node\_demand, total\_remaining\_demand
        \item Operators: AppendOperator, InsertOperator, SwapOperator, ReverseSegmentOperator, RelocateOperator, MergeRoutesOperator
    \end{itemize}
    \item \textbf{MKP}
    \begin{itemize}
        \item Data source: \url{https://people.brunel.ac.uk/$\sim$mastjjb/jeb/orlib/files/}
        \item Train instances: 20 generated instances
        \item Validation instances: gmknap1\_1$\sim$mknap1\_7
        \item Test instances: mknapcb1\_1$\sim$mknapcb1\_5, mknapcb4\_1$\sim$mknapcb4\_5
        \item Basic heuristics: block flip, greedy by cost benefit, greedy by density, greedy by least remaining capacity, greedy by profitto weight ratio, greedy by profit, greedy by resource balance, greedy by weight, greedy improvement, k flip, single swap heuristic, 2 opt
        \item Problem states: average\_profit, profit\_variance, average\_weight\_per\_resource, weight\_variance\_per\_resource, total\_weights, capacity\_to\_weight\_ratio, weights\_with\_epsilon, profit\_to\_weight\_ratio, solution\_density, average\_remaining\_capacity, remaining\_capacity\_variance, total\_remaining\_items, feasibility\_ratio, utilized\_capacity\_ratio, included\_items, included\_profits, item\_profitability\_in\_solution
        \item Operators: ToggleOperator, AddOperator, RemoveOperator, SwapOperator, FlipBlockOperator
    \end{itemize}
    \item \textbf{JSSP}
    \begin{itemize}
        \item Data source: \url{https://people.brunel.ac.uk/$\sim$mastjjb/jeb/orlib/files/}
        \item Train instances: 20 generated instances
        \item Validation instances: LA21$\sim$LA30
        \item Test instances: LA01$\sim$LA20
        \item Basic heuristics: first come first served, least work remaining, longest job next, longest processing time first, most work remaining, shift operator, shortest job next, shortest processing time first, 2opt, 3opt
        \item Problem states: average\_operation\_time, max\_operation\_time, min\_operation\_time, std\_deviation\_operation\_time, job\_operation\_time\_range, average\_job\_length, max\_job\_length, min\_job\_length, machine\_utilization, job\_diversity, num\_finished\_jobs, num\_unfinished\_jobs, average\_job\_completion, max\_job\_completion\_time, min\_job\_completion\_time, std\_dev\_job\_completion\_time, average\_machine\_completion, max\_machine\_completion\_time, min\_machine\_completion\_time, std\_dev\_machine\_completion\_time, average\_idle\_time\_per\_machine, proportion\_of\_finished\_jobs, proportion\_of\_unfinished\_jobs
        \item Operators: AdvanceOperator, SwapOperator, ReverseSequenceOperator, ShiftOperator
    \end{itemize}
    \item \textbf{MaxCut}
    \begin{itemize}
        \item Data source: \url{https://grafo.etsii.urjc.es/optsicom/maxcut.html}
        \item Train instances: 20 generated instances
        \item Validation instances: g11$\sim$g20
        \item Test instances: g1$\sim$g10
        \item Basic heuristics: balanced cut, greedy swap, highest delta edge, highest delta node, highest weight edge, most weight neighbors, multi swap 2, simulated annealing
        \item Problem states: average\_node\_degree, edge\_density, average\_edge\_weight, max\_edge\_weight, min\_edge\_weight, standard\_deviation\_edge\_weight, weighted\_degree\_distribution, imbalance\_ratio, cut\_value, average\_cut\_edge\_weight, selected\_nodes\_ratio, unselected\_nodes\_ratio, internal\_edges, edge\_weight\_variance\_within\_sets, boundary\_nodes, boundary\_node\_ratio
        \item Operators: InsertNodeOperator, InsertEdgeOperator, SwapOperator, DeleteOperator
    \end{itemize}
\end{itemize}

\section{Detailed Experiment Results}
\label{sec:detailed_experiment_results}
This appendix complements the main text with complete, per-instance results. We first report the performance achieved after heuristic evolution (Section~\ref{sec:detailed_evolution_results}); we then present the full outcomes of the problem solving stage when an LLM acts as the online selector (Section~\ref{sec:detailed_problem_solving_results_with_llm}). Unless otherwise noted, lower values indicate better solutions, and all gaps are expressed in percent relative to upper bound or best known.

\subsection{Detailed Evolution Results}
\label{sec:detailed_evolution_results}
The tables below list the optimality gaps obtained by the original seed heuristics and by their evolved counterparts generated by HeurAgenix. Because the public interfaces of ReEvo and EoH can evolve only the nearest-neighbor heuristic for TSP, comparisons with these two baselines are reported exclusively for that setting.

\begin{table}[H]
    \centering
    \caption{Per-instance optimality gaps (\%; lower is better) \textbf{before} and \textbf{after} heuristic evolution on all benchmark problems. ReEvo and EoH are included only for TSP–nearest-neighbor due to interface limitations. All heuristics are deterministic; therefore variances are zero.}
    \label{tab:evolution_result}
    \resizebox{\textwidth}{!}{
    \begin{tabular}{c|cc|cc|cccc}
    \toprule
    TSP & Cheapest insertion & Evolved & Farthest insertion & Evolved & Nearest neighbor & Evolved (Ours) & EoH & ReEvo \\
    kroA100 & 19.25 & 6.16 & 16.45 & 9.12 & 30.66 & 4.5 & 9.81 & 12.11 \\
    kroA150 & 16.27 & 8.22 & 14.04 & 8.34 & 31.69 & 10.62 & 16.12 & 11.33 \\
    kroB100 & 15.93 & 10.86 & 10.96 & 10.86 & 26.4 & 9.49 & 14.99 & 14.49 \\
    kroB200 & 22.3 & 11.81 & 17.73 & 11.81 & 14.76 & 10.2 & 16.05 & 16.02 \\
    kroC100 & 29.01 & 10.79 & 4.98 & 10.79 & 26.8 & 7.02 & 17.63 & 12.41 \\
    bier127 & 21.87 & 8.55 & 12.89 & 8.55 & 25.62 & 9.26 & 19.04 & 16.44 \\
    tsp225 & 18.12 & 7.48 & 14.49 & 7.48 & 28.35 & 5.15 & 17.86 & 13.4 \\
    a280 & 23.85 & 10.04 & 13.07 & 10.04 & 22.55 & 8.27 & 19.68 & 14.91 \\
    pcb442 & 22.31 & 10.81 & 18.86 & 10.81 & 22.02 & 12.92 & 15.42 & 16.42 \\
    gr666 & 17.7 & 9.41 & 19.19 & 9.41 & 24.67 & 12.63 & 16.8 & 19.74 \\
    pr152 & 11.28 & 6.47 & 6.6 & 2.41 & 16.31 & 4.36 & 16.55 & 14.83 \\
    pr1002 & 19.56 & 11.08 & 25.01 & 11.08 & 27.82 & 10.67 & 24.48 & 20.17 \\
    pr2392 & 21.7 & 15.02 & 28.82 & 15.02 & 21.99 & 12.68 & 18.49 & 24.93 \\
    \midrule
    average & 19.94 & 9.75 & 15.62 & 9.67 & 24.59 & 9.06 & 17.15 & 15.94 \\
    \midrule
    CVRP & Cheapest insertion & Evolved & Farthest insertion & Evolved & Nearest neighbor & Evolved \\
    A-n80-k10 & 20.6 & 18.67 & 29.57 & 24.12 & 33.26 & 20.6 \\
    B-n78-k10 & 42.92 & 22.6 & 36.94 & 22.93 & 43.98 & 42.92 \\
    E-n101-k14 & 39.93 & 25.21 & 85.1 & 25.02 & 55.39 & 39.93 \\
    F-n135-k7 & 41.82 & 45.87 & 23.84 & 22.34 & 54.22 & 41.82 \\
    M-n200-k17 & 36 & 24.94 & 104 & 36.55 & 56 & 36 \\
    P-n101-k4 & 38.03 & 16.89 & 30.1 & 24.38 & 49.93 & 38.03 \\
    \midrule
    average & 36.55 & 25.70 & 51.59 & 30.56 & 48.80 & 36.55 \\
    \midrule
    MKP & Greedy by profit & Evolved & Greedy by weight & Evolved & Greedy by density & Evolved \\
    mknapcb1\_1 & 20.63 & 3.39 & 27.23 & 7.71 & 7.71 & 4.05 \\
    mknapcb1\_2 & 18.07 & 3.43 & 33.44 & 4.59 & 4.59 & 1.01 \\
    mknapcb1\_3 & 21.31 & 4.3 & 34.62 & 4.11 & 4.11 & 0.98 \\
    mknapcb1\_4 & 13.51 & 6.19 & 28.75 & 16.6 & 16.6 & 3.72 \\
    mknapcb1\_5 & 13.58 & 4.82 & 27.99 & 8.19 & 8.19 & 2.76 \\
    mknapcb4\_1 & 14.13 & 6.22 & 30.01 & 6.57 & 6.57 & 2.23 \\
    mknapcb4\_2 & 23.9 & 4.67 & 31.02 & 6.12 & 6.12 & 4.38 \\
    mknapcb4\_3 & 21.88 & 6 & 26.25 & 9.89 & 9.89 & 2.48 \\
    mknapcb4\_4 & 9.66 & 7.58 & 25.85 & 8.96 & 8.96 & 3.05 \\
    mknapcb4\_5 & 10.75 & 3.61 & 30.99 & 7.55 & 7.55 & 2.22 \\
    \midrule
    average & 16.74 & 5.02 & 29.62 & 8.03 & 8.03 & 2.69 \\
    \midrule
    JSSP & Most work remaining & Evolved & First come first served & Evolved & Shortest processing time first & Evolved \\
    LA01 & 32.13 & 40.01 & 241.14 & 24.62 & 119.52 & 24.62 \\
    LA02 & 49.92 & 28.40 & 199.54 & 28.85 & 196.64 & 28.85 \\
    LA03 & 33.5 & 34.84 & 164.49 & 29.48 & 78.56 & 29.48 \\
    LA04 & 69.49 & 51.36 & 272.03 & 42.2 & 158.47 & 42.2 \\
    LA05 & 12.31 & 21.75 & 200 & 14.17 & 141.82 & 14.17 \\
    LA06 & 24.3 & 18.9 & 221.17 & 16.41 & 155.62 & 16.41 \\
    LA07 & 24.49 & 22.92 & 192.58 & 27.3 & 112.13 & 27.3 \\
    LA08 & 48.09 & 22.48 & 241.6 & 11.94 & 191.77 & 11.94 \\
    LA09 & 33.54 & 19.45 & 226.71 & 25.45 & 178.13 & 25.45 \\
    LA10 & 27.77 & 11.9 & 253.03 & 5.64 & 127.77 & 5.64 \\
    LA11 & 27 & 28.4 & 218.41 & 29.05 & 158.92 & 29.05 \\
    LA12 & 22.81 & 18.67 & 232.24 & 23.77 & 140.62 & 23.77 \\
    LA13 & 19.22 & 24.43 & 230 & 16.26 & 136.61 & 16.26 \\
    LA14 & 5.42 & 23.68 & 243.65 & 15.87 & 159.37 & 15.87 \\
    LA15 & 25.19 & 35.96 & 227.17 & 19.47 & 161.56 & 19.47 \\
    LA16 & 52.7 & 19.05 & 312.49 & 30.16 & 265.71 & 30.16 \\
    LA17 & 44.77 & 18.88 & 399.87 & 14.03 & 296.81 & 14.03 \\
    LA18 & 33.25 & 50.59 & 432.19 & 33.02 & 257.9 & 33.02 \\
    LA19 & 41.09 & 26.84 & 430.29 & 24.23 & 335.63 & 24.23 \\
    LA20 & 55.65 & 41.13 & 332.93 & 34.15 & 235.81 & 34.15 \\
    \midrule
    average & 34.13 & 27.99 & 263.58 & 23.30 & 180.47 & 23.3 \\
    \midrule
    MaxCut & Most weight neighbors & Evolved & Highest weight edge & Evolved & Balanced cut & Evolved \\
    g1 & 5.26 & 1.56 & 9.71 & 1.89 & 17.13 & 1.56 \\
    g2 & 4.74 & 1.72 & 10.34 & 1.57 & 16.77 & 1.72 \\
    g3 & 5.03 & 1.73 & 9.46 & 1.94 & 17.57 & 1.73 \\
    g4 & 5.67 & 2.25 & 9.38 & 1.89 & 17.91 & 2.25 \\
    g5 & 5.06 & 1.81 & 10.93 & 1.98 & 16.88 & 1.81 \\
    g6 & 29.48 & 11.52 & 55.05 & 9.14 & 100.69 & 11.52 \\
    g7 & 31.51 & 9.97 & 54.74 & 10.97 & 104.44 & 9.97 \\
    g8 & 32.05 & 10.17 & 58.92 & 12.11 & 112.16 & 10.17 \\
    g9 & 27.36 & 9.49 & 57.25 & 10.66 & 96.88 & 9.49 \\
    g10 & 26.15 & 14.3 & 59.7 & 9.15 & 103.7 & 14.3 \\
    \midrule
    average & 17.23 & 6.45 & 33.55 & 6.13 & 60.41 & 6.45 \\
    \bottomrule
    \end{tabular}}
\end{table}

\subsection{Detailed Problem Solving Results with LLM}
\label{sec:detailed_problem_solving_results_with_llm}
We next provide exhaustive results for the online-selection phase, where GPT-4o chooses among the evolved heuristic pool.  The same baselines as in Table \ref{tab:result_with_llm} of the main text are included for completeness.

\begin{table}[H]
    \center
    \caption{Per-instance optimality gaps (\%; lower is better) of all evaluated methods on the five problems. Bold indicates the \textbf{best result}, underline the \underline{second-best}. Values tagged with $^\dagger$ are taken from external sources; variance is omitted when it equals~0 or results are taken from external sources.}
    \label{tab:result_with_llm}
    \scriptsize
    \begin{tabular}{c|ccccccc}
    \toprule
    TSP & GLS & ACO & OR-Tools & EoH+GLS & ReEvo+GLS & ReEvo+ACO & Ours\\
    kroA100 & 2.98 $\pm$ 0.2 & 3.19 $\pm$ 3.1 & 6.35 $\pm$ 0.4 & \textbf{0.00} & \textbf{0.00} & \underline{2.73} $\pm$ 0.7 & \textbf{0.00}\\
    kroA150 & \underline{0.58} & 3.63 $\pm$ 3.3 & 4.74 $\pm$ 0.3 & \textbf{0.00} & \textbf{0.00} & 4.39 $\pm$ 2.8 & \textbf{0.00}\\
    kroB100 & 5.59 $\pm$ 0.8 & 4.58 $\pm$ 1.9 & 2.6 $\pm$ 0.1 & \textbf{0.00} & \textbf{0.00} & \underline{1.13} $\pm$ 0.3 & \textbf{0.00}\\
    kroB200 & 1.78 $\pm$ 0.8 & 2.91 $\pm$ 1.8 & 3.76 $\pm$ 1.9 & 0.38 $\pm$ 0.2 & \underline{0.18} & 1.06 $\pm$ 0.8 & \textbf{0.11}\\
    kroC100 & 7.1 $\pm$ 1.7 & 3.69 $\pm$ 1.9 & 12.74 $\pm$ 0.2 & \textbf{0.00} & \textbf{0.00} & \underline{3.58} $\pm$ 2.9 & \textbf{0.00}\\
    bier127 & 5.36 $\pm$ 1.5 & 5.68 $\pm$ 0.3 & 5.62 $\pm$ 0.4 & \underline{0.36} & \textbf{0.01} & 11.04 $\pm$ 3.5 & 1.29 $\pm$ 0.6\\
    tsp225 & 2.32 $\pm$ 0.1 & 5.05 $\pm$ 2.9 & 8.45 $\pm$ 0.7 & \underline{0.48} & 1.15 $\pm$ 0.3 & 1.7 $\pm$ 1.5 & \textbf{0.23} $\pm$ 0.1\\
    a280 & 6.44 $\pm$ 0.6 & 6.44 $\pm$ 3.9 & 12.54 $\pm$ 1.0 & \textbf{0.12} & \textbf{0.12} & 6.33 $\pm$ 2.2 & \underline{0.16} $\pm$ 0.1\\
    pcb442 & 3.15 $\pm$ 1.4 & 3.15 $\pm$ 2.7 & 7.94 $\pm$ 1.3 & 1.06 $\pm$ 0.1 & \underline{0.82} $\pm$ 0.6 & 2.94 $\pm$ 2.9 & \textbf{0.79} $\pm$ 0.2\\
    gr666 & 4.26 $\pm$ 1.3 & 4.26 $\pm$ 1.8 & 14.38 $\pm$ 0.2 & 2.65 $\pm$ 0.6 & \underline{1.08} $\pm$ 0.5 & 6.89 $\pm$ 2.0 & \textbf{0.93} $\pm$ 0.2\\
    pr152 & \underline{1.89} & 13.71 $\pm$ 2.2 & 2.92 $\pm$ 0.1 & 1.79 $\pm$ 0.1 & 1.89 & 9.31 $\pm$ 3.0 & \textbf{0.19} $\pm$ 0.1\\
    pr1002 & 4.96 $\pm$ 4.4 & 4.96 $\pm$ 2.4 & 22.18 $\pm$ 0.6 & 14.43 $\pm$ 4.9 & \underline{1.89} $\pm$ 0.6 & 17.96 $\pm$ 1.8 & \textbf{1.78} $\pm$ 0.6\\
    pr2392 & 4.74 $\pm$ 1.1 & 4.74 $\pm$ 3.9 & 8.45 $\pm$ 1.2 & \underline{3.87} $\pm$ 1.4 & 5.6 $\pm$ 1.4 & 6.37 $\pm$ 2.0 & \textbf{1.08} $\pm$ 0.5\\
    \midrule
    average & 3.93 & 5.08 & 8.67 & 1.93 & \underline{0.98} & 5.80 & \textbf{0.50} \\
    \midrule
    CVRP & OR-Tools & ACO & ReEvo+ACO & Ours\\
    A-n80-k10 & \underline{6.41} $\pm$ 0.3 & 34.96 $\pm$ 3.1 & 13.34 $\pm$ 2.9 & \textbf{3.97} $\pm$ 1.3\\
    B-n78-k10 & \underline{3.69} $\pm$ 0.2 & 0.20 $\pm$ 8.7 & 11.37 $\pm$ 5.9 & \textbf{2.54} $\pm$ 1.2\\
    E-n101-k14 & \underline{11.06} $\pm$ 0.5 & 49.30 $\pm$ 2.9 & 28.68 $\pm$ 5.8 & \textbf{10.97} $\pm$ 2.9\\
    F-n135-k7 & \underline{13.43} $\pm$ 1.3 & 50.17 $\pm$ 6.7 & 17.47 $\pm$ 5.5 & \textbf{8.52} $\pm$ 1.2\\
    M-n200-k17 & \underline{14.04} $\pm$ 0.6 & 64.16 $\pm$ 3.9 & 33.73 $\pm$ 6.6 & \textbf{9.88} $\pm$ 1.7\\
    P-n101-k4 & \underline{8.81} $\pm$ 0.8 & 59.18 $\pm$ 2.1 & 25.11 $\pm$ 8.4 & \textbf{3.08} $\pm$ 0.2\\
    \midrule
    average & \underline{9.57} & 43.00 & 21.62 & \textbf{6.49} \\
    \midrule
    MKP & QICSA($^\dagger$) & PSO($^\dagger$) & ACO & ReEvo+ACO & Ours\\
    mknapcb1\_1 & \underline{3.96} & 7.61 & 8.67 & 8.45 $\pm$ 2.8 & \textbf{0.33}\\
    mknapcb1\_2 & \underline{5.74} & 8.36 & 9.45 & 7.97 $\pm$ 1.8 & \textbf{0.20}\\
    mknapcb1\_3 & \underline{4.36} & 7.34 & 12.07 & 5.65 $\pm$ 0.7 & \textbf{0.15}\\
    mknapcb1\_4 & \underline{3.43} & 6.28 & 5.41 & 5.12 $\pm$ 1.8 & \textbf{1.23} $\pm$ 0.8\\
    mknapcb1\_5 & \underline{4.74} & 7.60 & 7.96 & 5.51 $\pm$ 1.0 & \textbf{0.72} $\pm$ 0.3\\
    mknapcb4\_1 & 5.50 & 9.40 & 16.05 & \underline{3.82} $\pm$ 0.8 & \textbf{0.55} $\pm$ 0.1\\
    mknapcb4\_2 & 6.37 & 9.38 & 10.39 & \underline{4.42} $\pm$ 1.3 & \textbf{1.05} $\pm$ 0.1\\
    mknapcb4\_3 & 6.51 & 9.37 & 12.12 & \underline{4.63} $\pm$ 0.9 & \textbf{0.62} $\pm$ 0.2\\
    mknapcb4\_4 & 6.13 & 8.19 & 10.59 & \underline{4.65} $\pm$ 0.6 & \textbf{1.03} $\pm$ 0.5\\
    mknapcb4\_5 & 5.85 & 9.23 & 14.89 & \underline{1.79} $\pm$ 0.1 & \textbf{0.91} $\pm$ 0.6\\
    \midrule
    average & 5.26 & 8.28 & 10.76 & \underline{5.20} & \textbf{0.68} \\
    \midrule
    JSSP & ACO($^\dagger$) & PSO($^\dagger$) & GWO($^\dagger$) & Ours\\
    LA01 & \textbf{0.00} & \textbf{0.00} & \textbf{0.00} & \textbf{0.00}\\
    LA02 & \textbf{0.00} & \textbf{0.00} & \textbf{0.00} & \textbf{0.00}\\
    LA03 & 4.36 & \underline{1.01} & \textbf{0.00} & \textbf{0.00}\\
    LA04 & \underline{3.56} & \underline{3.56} & \textbf{0.00} & \textbf{0.00}\\
    LA05 & \textbf{0.00} & \textbf{0.00} & \textbf{0.00} & \textbf{0.00}\\
    LA06 & \textbf{0.00} & \textbf{0.00} & \textbf{0.00} & \textbf{0.00}\\
    LA07 & \textbf{0.00} & \textbf{0.00} & \textbf{0.00} & \textbf{0.00}\\
    LA08 & \textbf{0.00} & \textbf{0.00} & \textbf{0.00} & \textbf{0.00}\\
    LA09 & \textbf{0.00} & \textbf{0.00} & \textbf{0.00} & \textbf{0.00}\\
    LA10 & \textbf{0.00} & \textbf{0.00} & \textbf{0.00} & \textbf{0.00}\\
    LA11 & \textbf{0.00} & \textbf{0.00} & \textbf{0.00} & \textbf{0.00}\\
    LA12 & \textbf{0.00} & \textbf{0.00} & \textbf{0.00} & \textbf{0.00}\\
    LA13 & \textbf{0.00} & \textbf{0.00} & \textbf{0.00} & \textbf{0.00}\\
    LA14 & \textbf{0.00} & \textbf{0.00} & \textbf{0.00} & \textbf{0.00}\\
    LA15 & \underline{0.41} & \textbf{0.00} & \textbf{0.00} & \textbf{0.00}\\
    LA16 & \textbf{0.00} & 6.35 & \underline{1.16} & 1.69 $\pm$ 0.2\\
    LA17 & \textbf{0.00} & 3.57 & \underline{0.77} & 1.79 $\pm$ 0.1\\
    LA18 & \textbf{0.00} & 4.36 & 1.30 & \underline{0.24} $\pm$ 0.1\\
    LA19 & \textbf{0.00} & 3.92 & \underline{0.36} & 1.19 $\pm$ 0.2\\
    LA20 & \textbf{0.00} & \underline{1.11} & 3.88 & \textbf{0.00}\\
    \midrule
    average & 0.42 & 1.19 & \underline{0.37} & \textbf{0.25} \\
    \midrule
    MaxCut & SS($^\dagger$) & CirCut($^\dagger$) & VNSPR($^\dagger$) & Ours\\
    g1 & \textbf{0.00} & \textbf{0.00} & \underline{0.03} & \textbf{0.00}\\
    g2 & \textbf{0.00} & 0.17 & \underline{0.04} & 0.07\\
    g3 & \textbf{0.00} & \underline{0.01} & \textbf{0.00} & 0.09\\
    g4 & \textbf{0.00} & \underline{0.11} & 0.39 & 0.44 $\pm$ 0.1\\
    g5 & \textbf{0.00} & \underline{0.01} & 0.28 & 0.22 $\pm$ 0.1\\
    g6 & \underline{0.60} & \textbf{0.00} & 3.49 & \textbf{0.00}\\
    g7 & \underline{1.20} & \textbf{0.00} & 4.99 & 1.30 $\pm$ 0.2\\
    g8 & 1.00 & \textbf{0.10} & 4.89 & \underline{0.80} $\pm$ 0.1\\
    g9 & \underline{0.68} & \textbf{0.10} & 2.73 & 2.73 $\pm$ 0.3\\
    g10 & \underline{0.35} & \textbf{0.20} & 4.50 & \underline{0.35} $\pm$ 0.1\\
    \midrule
    average & \underline{0.38} & \textbf{0.07} & 2.13  & 0.60 \\
    \bottomrule
    \end{tabular}
\end{table}

\end{document}